%% file: main.tex
\documentclass[10pt,twocolumn,letterpaper]{article}

\usepackage{cvpr}              % To produce the CAMERA-READY version
\input{preamble}

\definecolor{cvprblue}{rgb}{0.21,0.49,0.74}

\definecolor{color1}{rgb}{0.8, 1.0, 0.8} % Light pastel green  
\definecolor{color2}{rgb}{0.85, 0.95, 0.7} % Pastel green-yellow  
\definecolor{color3}{rgb}{0.95, 0.9, 0.6} % Light pastel yellow  
\definecolor{color4}{rgb}{1.0, 0.85, 0.5} % Soft pastel yellow-orange  
\definecolor{color5}{rgb}{1.0, 0.75, 0.6} % Light pastel orange  
\definecolor{color6}{rgb}{1.0, 0.65, 0.65} % Soft peachy pink  
\definecolor{color7}{rgb}{0.95, 0.6, 0.7} % Pastel pink  
\definecolor{color8}{rgb}{0.9, 0.5, 0.6} % Soft pastel red  pastel red
\usepackage[pagebackref,breaklinks,colorlinks,allcolors=cvprblue]{hyperref}

\def\paperID{5691} % *** Enter the Paper ID here 5691
\def\confName{CVPR}
\def\confYear{2025}

\newcommand{\method}{{Gen3DEval: Using vLLMs for Automatic Evaluation of Generated 3D Objects }\xspace}
\newcommand{\shortmethod}{Gen3DEval\xspace}

\title{\method}

\author{
Shalini Maiti\textsuperscript{*†}\quad 
Lourdes Agapito\textsuperscript{†}\quad 
Filippos Kokkinos\textsuperscript{*}\\
\vspace{0.1cm} \\
\textsuperscript{*}Meta AI\quad 
\textsuperscript{†}University College London
\vspace{0.2cm}
}

\begin{document}
\twocolumn[{%
\renewcommand\twocolumn[1][]{#1}%
\maketitle
\begin{center}
\centering
\captionsetup{type=figure}
% \vspace{-7.8cm}
% \vspace{-0.6cm}
\includegraphics [width=0.93\linewidth]{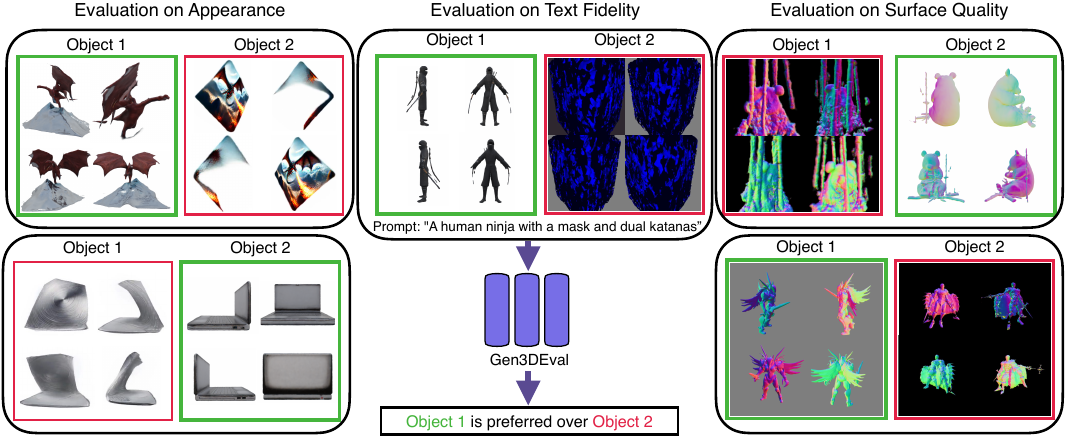}
%\vspace{-0.3cm}
\caption{\textbf{Gen3DEval}: A holistic ranking metric to assess the quality of generated 3D objects on appearance, surface quality and text fidelity using a vision large language model (vLLM) which is trained to choose the better out of two objects on the three evaluation dimensions (appearance, text fidelity or surface quality).}
\label{fig:teaser}
\end{center}%
}]

\input{sec/0_abstract}    
\input{sec/1_intro}
% \input{sec/2_formatting}
\input{sec/2_rw}
% \input{sec/3_finalcopy}
\input{sec/3_method}
\input{sec/4_experiment}
% \input{sec/5_datasets}
% \input{sec/6_training_details}
\input{sec/7_results}
\input{sec/8_conclusion}
\section*{Acknowledgements}
We thank Antoine Toisoul for help in curating the synthetic data for our experiments. 
Shalini Maiti has been supported by a sponsored research award from Meta.

{
    \small
    \bibliographystyle{ieeenat_fullname}
    \bibliography{main}
    % \bibliography{vedaldi_general}
    % \bibliography{vedaldi_specific}
}

\input{sec/X_suppl}

\end{document}

%% file: preamble.tex
\usepackage{multirow}
\usepackage{booktabs}
\usepackage{graphicx}
\usepackage{float}
\usepackage[dvipsnames,table]{xcolor}

\usepackage{tabularx}
\usepackage[accsupp]{axessibility}

%% file: sec/0_abstract.tex
\begin{abstract}

Rapid advancements in text-to-3D generation require robust and scalable evaluation metrics that align closely with human judgment, a need unmet by current metrics such as PSNR and CLIP, which require ground-truth data or focus only on prompt fidelity. To address this, we introduce \shortmethod, a novel evaluation framework that leverages vision large language models (vLLMs) specifically fine-tuned for 3D object quality assessment. \shortmethod evaluates text fidelity, appearance, and surface quality by analyzing 3D surface normals, without requiring ground-truth comparisons, bridging the gap between automated metrics and user preferences.
% Our benchmark, comprising 80 diverse prompts, sets a new standard for evaluating text-to-3D methods across multiple quality dimensions. 
Compared to state-of-the-art task-agnostic models, \shortmethod demonstrates superior performance in user-aligned evaluations, placing it as a comprehensive and accessible benchmark for future research on text-to-3D generation. The project page can be found here: \href{https://shalini-maiti.github.io/gen3deval.github.io/}{https://shalini-maiti.github.io/gen3deval.github.io/}.

\end{abstract}

%% file: sec/1_intro.tex
\section{Introduction}
The domain of text-to-3D generation has advanced significantly in recent years, driven by the rise of scalable architectures like diffusion models~\cite{LatentDiffusion}, neural radiance fields (NeRF)~\cite{Nerf}, and Gaussian splatting~\cite{kerbl3Dgaussians}. However, the field lacks standardized, human-aligned evaluation metrics that can reliably assess these assets and the methods that produce them. Existing metrics—such as CLIP~\cite{clip_radford} scores evaluate only limited aspects of the output like text fidelity and similarity-based measures like Peak-Signal-To-Noise Ratio (PSNR), SSIM~\cite{ssim}, Chamfer Distance, and Fréchet Inception Distance (FID)~\cite{fid} depend on ground-truth data making them inadequate and impractical for text-to-3D generation, where diverse outputs may correspond to a single prompt. In such cases, a unique, universally applicable reference does not exist, as multiple plausible 3D outputs can vary widely in style, appearance, and fidelity to the text. Meanwhile, FID computes a distributional similarity, which poses other challenges. Currently, there is no standardized large-scale validation set to serve as the ground-truth distribution for 3D assets, making FID computation difficult and inconsistent. Moreover, generating sufficient 3D assets to estimate this distribution requires significant computational resources, casting FID as an expensive and time-intensive metric. As a result, these metrics fall short of capturing the nuanced requirements of evaluating text-to-3D generation, where a scalable, human-aligned approach is essential. 
% for reliable evaluation.

While prior work such as GPT4VEval~\cite{Gpt4vEval} has leveraged GPT-4V~\cite{GPT4} for assessing 3D asset quality, GPT-4V is a general-purpose model not specifically trained for 3D quality assessment, which limits its effectiveness in this domain. Furthermore, it can be costly to deploy at scale, and in our experiments, we found that GPT-4o (which is the successor to GPT-4V) performed significantly worse than our method in aligning with human judgments of 3D asset quality.

To bridge this gap, we introduce \shortmethod, a vision-based large language model (vLLM) framework specifically fine-tuned to evaluate text-to-3D generation outputs in alignment with human preferences. Unlike existing metrics~\cite{clip_radford,ssim,lpips}, \shortmethod assesses not only text fidelity but also appearance and surface quality by analyzing rendered multi-view images. Supporting up to eight images as input, \shortmethod enables comprehensive assessment by leveraging multi-view renderings, such as RGB and normal maps, of generated 3D objects. Using multi-view images as input allows for compatibility across diverse 3D representations~\cite{Nerf, kerbl3Dgaussians,sdf}.

Built upon the recent vLLM early fusion approaches~\cite{llava,qwen, blip,fuyu}, our method processes these input renderings by first encoding each image through an image encoder, which translates them into visual tokens. These tokens are then integrated with text tokens and fed into a Llama3 model, allowing \shortmethod to interpret both visual and textual features of the 3D objects holistically. To ensure robust performance, we curate data from human assessments and further enhance our training dataset with synthetically generated perturbations of artist-created 3D objects, incorporating artifacts like floaters, transparency errors, text fidelity inconsistencies, excessively smooth surfaces etc.

A key component of our framework is \shortmethod-Bench, a benchmark dataset designed to standardize text-to-3D evaluations across various quality dimensions. Comprising 80 diverse prompts, \shortmethod-Bench facilitates consistent, human-aligned assessments of visual fidelity and aesthetic preferences. Our evaluation pipeline involves two main stages: first, it performs pairwise comparisons of 3D objects using multi-view renderings. Then it applies ELO rating metrics~\cite{elo_rating} to generate scores that closely align with human judgment. This process ensures robust and reliable evaluations across a broad range of 3D generative methods. To sum up, our contributions are: \begin{itemize} \item A state-of-the-art holistic evaluation method for text-to-3D generation that ranks methods across appearance, surface quality and text fidelity. \item A vLLM fine-tuned on the Llama3~\cite{llama3} model, using a synthetic dataset curated to reflect human preferences for evaluating generated 3D assets. \item A benchmark dataset, \shortmethod-Bench, comprising 80 prompts for ranking existing and future text-to-3D generation methods in a standardized manner. \end{itemize}

%% file: sec/2_rw.tex
% Related Work:
% Academic siblings of our work, i.e. alternative attempts in literature at trying to solve the same problem. 
% Goal is to “Compare and contrast” - how does their approach differ in either assumptions or method? If their method is applicable to our Problem Setting I expect a comparison in the experimental section. If not, there needs to be a clear statement why a given method is not applicable. 
% Note: Just describing what another paper is doing is not enough. We need to compare and contrast.

\input{figures/model_details}
\section{Related Work}

\textbf{Text-to-3D and Image-to-3D generation.}
In recent times, the landscape of text-to-3D generation has seen rapid growth with the advent of representations such as Neural Radiance Fields \cite{Nerf}, occupancy fields ~\cite{occupancy_fields}, SDF~\cite{sdf} and Gaussian Splats~\cite{kerbl3Dgaussians}, and the availability of large, publicly available datasets such as Objaverse ~\cite{objaverse,objaverseXL}.
Some of the earlier methods in the space currently include ~\cite{dreamfusion, assetgen, magic3d, DreamTimeAI, Magic123, dreamgaussian, scorejacobianchaining, prolific_dreamer, gaussiandreamer, HIFAHT} that optimizes a randomly-initialized 3D model via gradient descent conditioned on sampled outputs of a text-to-image generation model. 
Another direction of work include methods such as~\cite{mvdream, imagedream, zero123plus, syncdreamer, viewdiff, cat3d, nvist} that use multi-view diffusion models to fine-tune text-to-image models to quickly generate highly consistent multiple views or videos simultaneously from a single input image. 
Notably, another family of methods ~\cite{lrm, vfusion3d, assetgen} learns 3D priors from a large amount of data and a scalable architecture to directly output robust 3D outputs from text or image inputs.
With such an impressive pace of growth in this research domain, it is imperative for the presence of robust evaluation metrics and benchmarks to ensure continued progress in this field, which is a gap that \shortmethod attempts to bridge.

\textbf{3D Evaluation Metrics and benchmarks}
Classical 3D metrics like PSNR, Chamfer Distance, LPIPS~\cite{lpips} and SSIM~\cite{ssim} were developed to measure the quality of a generated 3D asset against a ground-truth asset. However, these are similarity metrics and measure the distance between generated and the ground-truth data. This is infeasible since a single text prompt can be mapped to many generated 3D outputs, with their quality or fidelity being independent of their similarity to any single generated asset.
% A similar case can be made for generating either 3D outputs or geometrically consistent video outputs from a single image since we only obtain a single view of an object to utilize as ground-truth, which is insufficient. 
We propose that instead of measuring similarity, we need to inject 3D quality priors into the method itself for the purpose of evaluation, which is the foundation of \shortmethod. 

Other metrics such as CLIP~\cite{clip_radford} scores have tried to measure alignment with text by using a standard benchmark of textual prompts and computing a corresponding score. However, they underwhelm with an increase in diversity of prompts. Another inadequacy is that only text-alignment is measured and not appearance or surface quality. \shortmethod addresses both of these aspects.
The work closest to ours in attempting to solve this problem is GPT-4V(ision) is a Human-Aligned Evaluator for Text-to-3D Generation~\cite{Gpt4vEval}. However, it uses GPT4~\cite{GPT4} off the shelf, which is a task-agnostic vLLM trained on half a trillion parameters whose API and checkpoints are not publicly available, making it costly to scale, whereas \shortmethod has been \textit{specifically} trained to evaluate text-to-3D objects, on 8.35 billion parameters and will be made available for public usage. Moreover, in our experiments, we found that \shortmethod performed significantly better than GPT4-o in aligning with human judgments of 3D asset quality. 

In the absence of quantifiable metrics, user studies have been popularly employed as the gold standard to evaluate 3D generation methods. However, this is time-consuming, cost-ineffective and lacks a standard procedure. Certain benchmarks such as T$^3$Bench~\cite{t3bench} and Dreamfusion prompts~\cite{dreamfusion} have been created to reduce the lack of standardization in this process. \shortmethod takes a step further in this direction by curating a benchmark \shortmethod-Bench with diverse prompts in terms of types objects, length and compositionality.

\textbf{Large Multi-modal Models}
The past couple of years has seen great strides made in the development of Large languages models (LLMs) like Llama~\cite{llama}, GPT-4~\cite{GPT4}, Claude~\cite{Claude}, Gemini~\cite{gemini},  and consequently, led to development of vLLMs such as LLaVA~\cite{llava}, BLIP~\cite{blip}, FUYU~\cite{fuyu} and more~\cite{llama3, GPT4, Claude}. They are powerful multi-modal models that display strong image and language reasoning. However, since these are general purpose models, they do not perform well on evaluating generated 3D objects. ~\cite{Gpt4vEval} showcases capabilities of GPT4 ~\cite{GPT4} to be able to align with human preference for the assessment of 3D objects. \shortmethod takes this effort further by fine-tuning a Llama3~\cite{llama3} model using a curated synthetic dataset for the specific purpose of introducing 3D aesthetic preference into the vision-language space and transforming that into a ready-to-use evaluation ranking metric.

 % gemini citation throws an error so it's been removed

%% file: figures/model_details.tex
% \begin{figure*}[!htbp]
%     \centering
%     % Adjust the height (e.g., 5cm) as needed, width is set to text width
%     \fbox{\rule{0pt}{5cm} \rule{\textwidth}{0pt}}
%     \caption{Full-width placeholder for a figure}
%     \label{fig:teaser_fig}
% \end{figure*}
% \vspace{5mm}
\begin{figure*}[tbh]
%\vspace{-9.5cm}
\begin{center}
\includegraphics [height=0.5\linewidth]{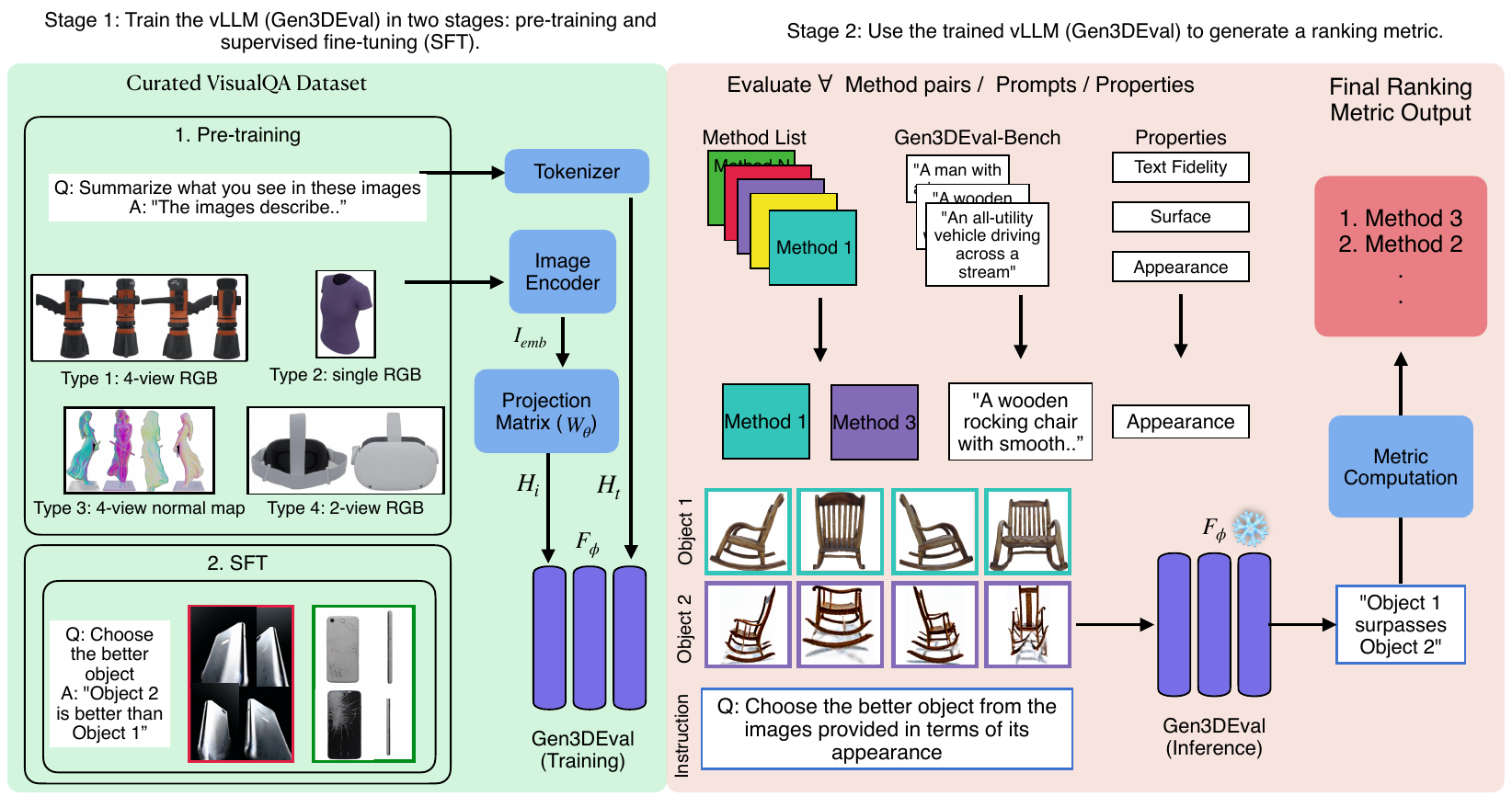}
\vspace{-0.6cm}
\end{center}
\caption{\textbf{\shortmethod framework:} In stage 1, we train a vLLM to choose which object is better in terms of appearance, surface quality or text fidelity. This is further divided into 2 parts. In pre-training, we train the vision-to-language projector using image summary VQA. In the supervised fine-tuning (SFT) stage, we use comparison data to train for instruction following and preference evaluation. In stage 2, we compute a ranking metric for the set of methods by applying the trained vLLM from stage 1 pairwise on \shortmethod-Bench prompts.}
% \vspace{-2mm}
\label{fig:model_details}
\end{figure*}

%% file: sec/3_method.tex
%\input{figures/model_details}
\section{Method}
The proposed method, \shortmethod is a vision-based large language model (vLLM) that interprets and assesses the quality of 3D generated objects. We train \shortmethod using a carefully curated Visual-Question-Answering (VQA) dataset, as detailed in Section \ref{sec:dataset}. This training enables \shortmethod to learn associations between visual cues in multi-view images and quality indicators such as text fidelity, surface detail, and appearance. We use up to eight multi-view RGB images—renderings that include RGB and normal maps—to capture comprehensive object details from multiple angles; see Figure \ref{fig:model_details} for an overview.

\subsection{Model Details}
The \shortmethod training process builds upon the LLaVA architecture~\cite{llava} and is organized into two sequential phases: pre-training and supervised fine-tuning. In the initial phase, each image is processed by an image encoder to produce visual embeddings $I_{emb}$. These embeddings are transformed into language-compatible tokens $H_{i}$ via a linear projection matrix $W_{\theta}$, enabling them to integrate seamlessly with the language-based representations. At the same time, a language tokenizer converts the natural language Question-Answer (QA) pairs into text tokens $H_{t}$. The model $F_{\phi}$ then receives both the image and text tokens as input, learning to predict the next token $H_{y}$ by maximizing the likelihood of the correct token.

For the image encoder, we initially consider CLIP~\cite{clip_radford}, which is commonly used across various vision-language models~\cite{llava, dreamfusion, imagen}. However, since the input images are rendered views of 3D objects, we also evaluate two additional encoders: DinoV2~\cite{dinov2} and Fit3D~\cite{fit3d}. DinoV2 generates visually consistent embeddings, while Fit3D is specifically designed to encode 2D images into features consistent with the underlying 3D scenes, making it particularly suited to our task. A comparison of these feature encoders is provided in Table~\ref{tbl:method_comparison} and discussed in Section~\ref{sec:image_encoders}.

During the pre-training phase, the weights of both the LLM and the image encoder are frozen, and only the weights of the linear projection matrix $W_{\theta}$ are updated. This selective tuning establishes alignment between the visual embeddings and language tokens, forming a foundation for integrated visual-language comprehension.

In the fine-tuning phase, we unfreeze both the projection matrix and the LLM, allowing them to be fine-tuned jointly, while keeping the image encoder’s weights frozen. This stage further specializes the model for 3D quality assessment, enhancing its sensitivity to features such as surface texture, text fidelity, and overall visual coherence across various prompts and multi-view renderings.

\subsection{Multi-view Input}
To effectively evaluate the quality of generated 3D objects, \shortmethod leverages multi-view input, using up to 8 rendered images uniformly panning each object. This approach is essential for capturing the complete appearance and surface consistency of 3D objects, as single-view images may overlook aspects like hidden surfaces and occlusions that become visible when observed from multiple viewpoints.

In pre-training, the input images range from a set of 1 to 4 RGB images or rendered surface normals panning the object in a $360^{\circ}$ round-table manner alongside a short summary in a QA pair. In fine-tuning, we input two sets of images, for object 1 and object 2. Each set consists upto 4 multi-view images each, therefore training a VQA sample consists upto 8 images and QA capturing the preference for the preferred object. The number of tokens per image is 576 and approx. 250 for the text of the question.

\input{figures/train_data_samples}
\section{\label{sec:training}Training Details}

The following section provides an in-depth look at the \shortmethod dataset, outlining its composition, structure, and the methodologies used to create a diverse and robust training set. We describe the dataset’s sources and organization, the approach to rendering multi-view images, and the use of human judgment and synthetic perturbations to enhance model alignment with 3D quality standards. Each component of the dataset is designed to support the pre-training and fine-tuning stages, ensuring comprehensive coverage of key attributes like appearance, surface consistency, and text fidelity.

\subsection{\label{sec:dataset}Dataset}

\shortmethod's training dataset is designed to train and evaluate the model’s ability to assess 3D object quality across various dimensions, including appearance, surface consistency, and text fidelity. It comprises three subsets a) 3D artist-created meshes, b) human preference data on generative method outputs and c) synthetic 3D comparison data. 

\textbf{3D Meshes:}
Comprising 140,000 high-quality 3D meshes created by artists, this internal dataset spans diverse semantic categories and provides a robust foundation for generalizing to different types of 3D content. Each asset comes with an accompanying text caption generated with Llama3.2~\cite{llama3}. We render each 3D asset from multiple viewpoints, creating three types of visual inputs: RGB images, alpha masks, and surface normals. These multi-view renderings allow \shortmethod to capture comprehensive visual cues necessary for accurate 3D evaluation.

\textbf{Human Annotations:}
To account for the nuances and irregularities inherent in 3D generative methods, we conducted a large-scale human preference study, collecting over 5K comparative data points across 13 different 3D generative methods~\cite{meshy, assetgen,texturegen, vfusion3d, flex3d, im3d, genie, openlrm, triposr, syncmvd, mvdream,imagedream,zero12345++}. Annotators viewed 360$^\circ$ videos of two 3D assets side-by-side and selected the preferred asset based on appearance and alignment with the corresponding generation prompt. 
% Check this out later
% 5,000
% We also added 15K samples of user-annotated images from various image-generation methods~\cite{Pick-a-Pic}.

This preference data was incorporated into the \shortmethod training dataset, enhancing the model’s alignment with human aesthetics and text fidelity expectations. We conducted an in-depth analysis of these annotations to identify common artifacts in modern text-to-3D generation methods, including (1) disconnected components, (2) Janus artifacts, (3) opacity inconsistencies, (4) floating elements, (5) overly smooth or irregular surfaces, (6) texture seams. Based on these insights, we replicated these artifacts in the 3D mesh data to scale up our dataset and further improve \shortmethod’s performance. Examples of these artifacts are illustrated in the appendix. 

% [~\ref{fig:suppl_pre_1},~\ref{fig:suppl_pre_2},~\ref{fig:suppl_sft_1},~\ref{fig:suppl_sft_2}].
 % Figure~\ref{}.

\textbf{Synthetic Data:}
To expand the training dataset, we applied controlled perturbations using Blender, NeRF, and Gaussian splatting techniques, simulating common artifacts and misalignments found in text-to-3D generative outputs. For 3D meshes, we introduced perturbations such as Laplacian smoothing~\cite{laplacian_smoothing}, beveling, random surface extrusions, and texture map alterations like blurring and seam introduction. Additionally, we fitted NeRF and Gaussian splats to the renderings of the artist-created 3D assets, providing a broader foundation for synthetic training data.

To further enrich this dataset, we introduced additional perturbations to the NeRF and Gaussian splatting by manually adding transparency artifacts, floating elements, and disconnected components, mimicking frequent issues observed in text-to-3D generative methods.

% Check this out later
To scale the text fidelity comparison dataset, we used the multi-view video diffusion model from IM-3D~\cite{im3d} and trellis~\cite{trellis} to generate single and multiple views of 3D objects with varied captions, focusing on changes to appearance attributes and composition of the objects. Textual perturbations were created with Llama3.2~\cite{llama3} by prompting the model to modify the original captions, introducing subtle variations. This approach allowed us to generate a large, diverse synthetic dataset tailored for text fidelity comparison. To ensure high relevance and quality, we applied CLIP~\cite{clip_radford} to filter out examples with low image-text similarity, resulting in a refined synthetic dataset for evaluating text fidelity. Figure \ref{fig:train_samples} showcases some samples of our SFT dataset.

\subsection{\label{sec:image_encoders}Image Encoders}
We evaluated 3 image encoders: CLIP~\cite{clip_radford}, DinoV2~\cite{dinov2}, Fit3D~\cite{fit3d} and the combinations of CLIP with DinoV2 and CLIP with Fit3D; reshaping them to match the bigger of the two and adding the values. For CLIP embeddings, the effective resolution is 336x336 pixels and for both Dinov2 and Fit3D, which internally fine-tune the Dinov2 base model to introduce 3D awareness to image features, the model uses a ViT~\cite{vit} backbone of patch size 14, effective resolution of 224x224 and embedding dimension 768. The results of these ablation experiments are reported in Table ~\ref{tbl:method_comparison}.

We observe that \shortmethod performs equally well on synthetic surface assessment in all the configurations with lowest accuracy score of 0.89 in the case of \shortmethod w/ Fit3D whereas on user-annotated out-of-domain (OOD) benchmark, CLIP clearly outperforms the rest. On the synthetic appearance benchmark, described in Section \ref{sec:synthetic_eval_dataset}, \shortmethod w/ Fit3D reports the lowest accuracy of 0.8, with the rest between .85-.89. On the other hand, on the human evaluation appearance dataset (in-domain methods) described in Section \ref{sec:human_eval_dataset}, \shortmethod w/ CLIP as well as a combination of Fit3D and CLIP report the best accuracy score of 0.9, followed by CLIP and DinoV2 (0.86), then Fit3D (0.81) and finally DinoV2 (0.77). In terms of generalization with OOD benchmarks, \shortmethod with CLIP outperforms the rest by a large margin.

On text fidelity benchmarks,  in keeping with our earlier observation, the performance on OOD text fidelity benchmark is much better for \shortmethod with CLIP (0.86) alone compared with the rest, with the nearest neighbour in CLIP and DinoV2 (0.74). Finally, on the synthetic text fidelity benchmark, \shortmethod w/ DinoV2 alone underwhelms reporting 0.75. While standalone numbers for Fit3D is better, CLIP reports high scores by itself as well as in conjunction with Fit3D and DinoV2.

Overall, we noticed that \shortmethod with CLIP embedding consistently performs well across all evaluation dimensions. As a result of this, \shortmethod uses CLIP encoder to extract image embeddings.

\subsection{Stage 1: Pre-training}
The objective of this training stage is to train the projection matrix to learn correlations between the image encoding space and the language description space. To train the projection matrix, we use the renderings of the 141,000 3D artist-created meshes and their accompanying text prompts.  We sample 40K single view image, 40K two-view images, 40K four-view images,  and 10K four-view rendered surface normal images and their corresponding captions. We also combine 11.4K of the four-view images mentioned above and combine them to form an image grid. In the case of surface normals, we process the captions to remove any aspect of appearance mentioned in them which is irrelevant to rendered normals.  All multi-view images are sampled uniformly from a $360^{\circ}$ azimuth with a fixed elevation angle. 

The training process involved a batch size of 16, learning rate of 1e$^{-3}$, a cosine learning rate scheduler with a warm-up ratio of 0.03, using the ADAM optimizer. We optimized the model using maximum likelihood for the next token prediction. Pre-training was conducted on 8 A100 GPUs over a period of 1 day, encompassing 8K iterations.

\subsection{Stage 2: Supervised Fine-tuning}
The objective of the supervised fine-tuning (SFT) stage is to jointly train the instruction-following large language model (LLM) and the pre-trained projection matrix for the task of selecting the best 3D object out of two based on text fidelity, 3D appearance and surface quality. For fine-tuning, we utilize the human-annotated data and the synthetically generated comparison data. The SFT dataset distribution is displayed in the appendix.

The fine-tuning process involved a batch size of 4, a learning rate of 2e$^{-6}$ for the projector, and a learning rate of 1e$^{-5}$ for the vLLM, with a cosine learning rate scheduler and a warm-up ratio of 0.03, using the ADAM optimizer. We optimized the model by using maximum likelihood of next token prediction. This stage was trained on 16 A100 GPUs for 18 hours, for 4K iterations.

%% file: figures/train_data_samples.tex
\begin{figure*}[tb]
% \vspace{-9cm}
\begin{center}
\includegraphics[width=0.92\linewidth]{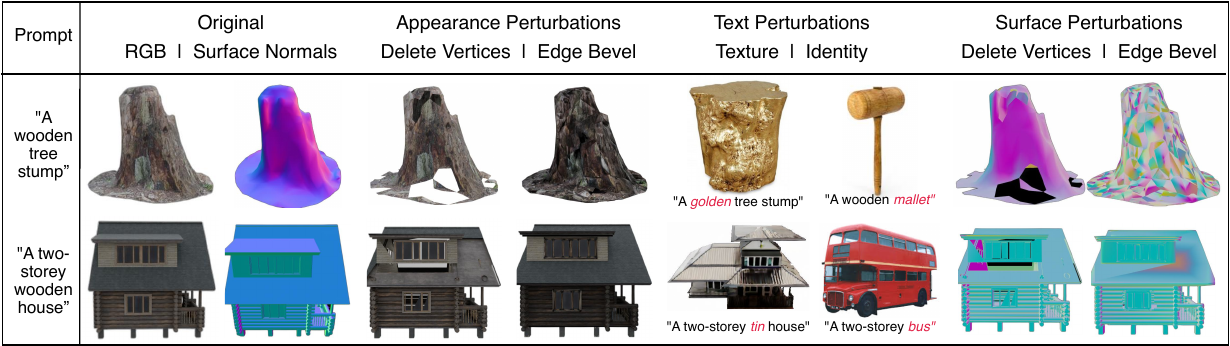}
\end{center}
%\vspace{-0.5cm}
\caption{\textbf{Training Dataset} We use single and multi-view RGB and surface normals renderings of a 3D object generated from a prompt. We take these objects and perturb them to simulate common appearance, surface and text-related artefacts in generative 3D methods.}

\label{fig:train_samples}
\end{figure*}

%% file: sec/4_experiment.tex
\input{figures/qualitative_comparison_fig}
\section{Experiments}

% You need to say here that you evaluatate 1) app. 2) Text faith 3) Surface. To do this you construct datasets using human annot and synthetic.  No repetitions with what was mentioned above. Give numbers and after that explain what you see in the table you made for all these datasets.  In the text below you keep mentioning the same things over and over. 

We evaluated the performance of \shortmethod using three distinct datasets. Their details are explained in the following subsections. We also performed ablation studies to explore the impact of different image encoders on the performance of \shortmethod. This involved varying the types of image encoders and delineating their respective contributions to the overall performance of the model in Section \ref{sec:image_encoders}.

\subsection{\label{sec:human_eval_dataset}Human Evaluation Dataset}
The purpose of the human evaluation dataset is to assess the alignment of \shortmethod with human preference on text fidelity, appearance and surface quality. We curated this dataset in the same way as we curated the human preference data for the supervised fine-tuning stage of training in Section \ref{sec:dataset} under Human Annotations. Post curation and processing, we split the total data by holding out $10\%$ of the total prompts (404) for the creation of this dataset and using the rest ($90\%$) for the supervised fine-tuning stage of training. This dataset has 506 VQA comparison data samples for a total of 40 prompts, annotated on the basis of appearance.
We also added 3 evaluation datasets, one each for appearance, surface and text fidelity generated from annotating pairwise evaluation and removing any ambiguities from methods as well as prompts that were not used as part of the training data. We use these to calculate out-of-domain (OOD) generalization performance of our \shortmethod.
% This dataset comprises pairwise asset comparison data collected from trials involving a set of text-to-3D generation models. The prompts used in this dataset are sourced from \cite{dreamfusion}, totaling 404 prompts. We further curated this dataset by including only those samples where all evaluators (three in total) unanimously selected the same 3D asset. We split this dataset at the prompt level, reserving 10\% of the prompts for the evaluation dataset.

\subsection{\label{sec:synthetic_eval_dataset}Synthetic Dataset}
The second dataset is a synthetic evaluation dataset that includes objects generated by 3D artists along with their synthetically perturbed counterparts, enabling controlled experimentation in a low-noise environment. We created and processed this dataset in the same way as we processed the synthetic dataset for appearance, surface and textual perturbations detailed in Section \ref{sec:dataset} under Synthetic Data. Given that the VQA training is so diverse, we further ensured no overlap with training captions and objects by filtering the evaluation dataset using a sentence similarity threshold using sentence transformer embeddings~\cite{SBERT}.
We also curated a portion of the synthetic evaluation dataset by applying surface perturbations to artist-drawn meshes and rendering the surface normals of these pairs of objects. All the filtering mechanisms were similarly applied to this dataset.

% introducing perturbations across various dimensions and tasked the model with choosing between two objects. The dimensions of perturbation include textual prompts, surface, textures, opacity, and color. We filtered this dataset to remove noisy data by applying similarity thresholds on images, text, and scale.

\subsection{\label{sec:benchmark}Benchmark Details}
We create a diverse set of 80 prompts, \shortmethod-Bench, which considers the diversity of objects, textures, and levels of composition. We determine the size of this benchmark with the consideration that text-to-3D generation is a time- and computation-intensive process, aiming to make the benchmark easily accessible. It is split between 40 animate (humanoids, animals) objects and 40 inanimate objects, as well as into 43 single object and 37 composite object prompts, i.e., combining multiple objects. The average number of words per prompt is 12.863. Refer to the supplementary for comparison with other prompt benchmarks.

\subsection{\label{sec:metric}Metric Computation}
To compute the metric, \shortmethod compares two methods at a time using the following procedure. First, it samples four images from a 360$^\circ$ RGB or surface normal video circling the object, at equal intervals covering a 360$^\circ$ view of the 3D asset per prompt per method for a pair of methods. 
% As input, it takes two sets of four multi-view images generated from the same prompt, along with a QA prompt asking the model to choose the better asset in terms of appearance, surface quality or text fidelity. 
For each prompt in \shortmethod-Bench and each pair of methods, \shortmethod is applied to a pair of assets at a time. It takes 8 input images (4 per object) alongwith the relevant QA prompt and parses the natural language output using \cite{llama3} to determine which 3D asset is better. Subsequently, it applies the ELO rating system to the parsed outputs and extracts an overall ranking metric. The generation prompt is only provided in the case of evaluation on text fidelity and not for appearance and surface. We treat them as separate tasks which enables us to compare any two generated assets, irrespective of the generated prompts. It also allows us to evaluate image-to-3D methods more effectively.

%% file: figures/qualitative_comparison_fig.tex
\begin{figure*}[tb]
% \vspace{-9cm}
\begin{center}
\includegraphics[width=0.9\linewidth]{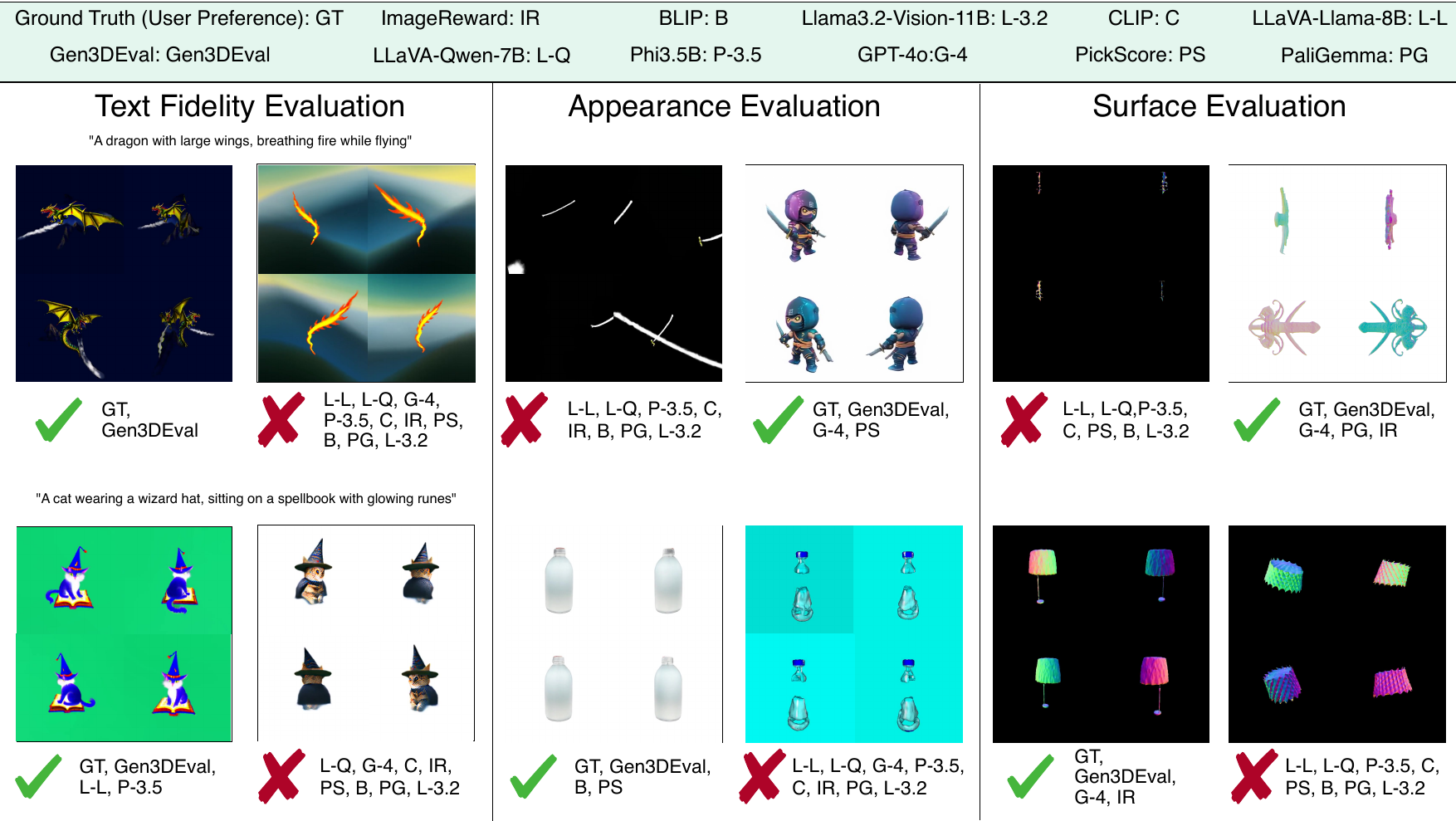}
% \vspace{-3cm}
\end{center}
\vspace{-0.45cm}
\caption{\textbf{Qualitative Comparison} of methods on samples of the evaluation dataset across text fidelity, appearance and surface evaluation.}

\label{fig:qualitative_method_comparsion}
\end{figure*}

%% file: sec/7_results.tex
% Results and Discussion:
% Shows the results of running Method on our problem described in Experimental Setup. Compares to baselines mentioned in Related Work. Includes statistics and confidence intervals. Includes statements on hyperparameters and other potential issues of fairness. Includes ablation studies to show that specific parts of the method are relevant. Discusses limitations of the method. 
%\input{figures/qualitative_comparison_fig}
\section{Results}

\subsection{\shortmethod and other evaluator methods }
We compare \shortmethod against classical baseline metrics such as CLIP as well state-of-the-art vLLMs on evaluation datasets described in Section \ref{sec:human_eval_dataset} and Section \ref{sec:synthetic_eval_dataset}.
We show that the model outperforms all the current methods on assessing appearance preference by a large margin on synthetic, user preference and out-of-domain evaluation data, demonstrating a strong correlation with human preference in the context of text-to-3D asset generation. In terms of text fidelity, \shortmethod outperforms CLIP~\cite{clip_radford}, which is the most popular metric for text fidelity evaluation. \shortmethod also narrowly outperforms ImageReward~\cite{imagereward} and PickScore~\cite{Pick-a-Pic} on the out-of-domain benchmark which has been curated to remove any ambiguous samples. We also outperform our baselines of surface comparisons data using only synthetically perturbed surface normals.

\input{tables/method_comparison_table}

Moreover, \shortmethod is the first method that unifies text-to-3D generation metrics by incorporating appearance as well as text fidelity metrics in a holistic manner as evidenced in Table~\ref{tbl:method_comparison}. We also provide a qualitative comparison of \shortmethod with other methods on a few samples from the evaluation dataset in Figure \ref{fig:qualitative_method_comparsion}.

We also report results for ablation studies for choice of image encoders used for the instruction tuning stage in Table ~\ref{tbl:method_comparison}.
% ~\ref{tbl:method_ablations}.
% \input{tables/method_ablations}
We note from the ablations that CLIP embeddings have a more consistent performance across all dimensions for the purpose of our metric. 

Finally, we note that while user studies or user preference data is the current gold standard, it can be noisy and uncorrelated. For instance, when it comes to text fidelity comparison, sometimes, the preference is influenced by appearance. Moreover, with short or simple generation prompts, it is difficult to pick one over the other. In case of appearance comparison, sometimes, the background or scale can influence our choice.

\subsection{Generative 3D Methods on \shortmethod-Bench}
Table \ref{tbl:leaderboard_methods} notes the results of \shortmethod on \shortmethod-Bench for a collection of 10 generative 3D methods, namely, Trellis~\cite{trellis}, AssetGen~\cite{assetgen}, Fantasia3D~\cite{fantasia3d}, TripoSR~\cite{triposr}, Magic123~\cite{Magic123},
Magic3d~\cite{magic3d}, Vfusion-3d~\cite{vfusion3d}, Dreamfusion~\cite{dreamfusion}, LatentNerf~\cite{latentnerf} and Flex3D~\cite{flex3d}. We rank them in the order of their performance for appearance, surface quality and text fidelity. The overall ranking is an average of the scores of appearance and text fidelity.

\input{tables/leaderboard_methods}

\subsection{Limitations}
\shortmethod's assessment of objects with janus can be slightly erratic. There is room for improvement for out-of-domain performance for surface evaluation, because of limited availability of diverse, annotated surface comparison data. We also note that for image-to-3D generation methods, the performance of methods are inherently influenced by a text-to-image generation pipeline. Therefore, composing a strong and consistent image benchmark would be a logical next step. While \shortmethod can be used for the purpose of evaluating any given pair of generated 3D objects, for the purpose of being used as a standard evaluation metric, comprehensive application across numerous methods is relevant to its performance; examples in the appendix.

% Figure ~\ref{fig:supp_limitations}.

%% file: tables/method_comparison_table.tex
\begin{table*}[tbh]
\centering
% \noindent
\resizebox{0.9\textwidth}{!}{\begin{tabular}{lcccccccc}
    &\multicolumn{3}{c}{Appearance} & \multicolumn{2}{c}{Surface} & \multicolumn{2}{c}{T-Fidelity}\\
    \cmidrule(lr){2-4} \cmidrule(lr){5-6} \cmidrule(lr){7-9}  
    & {Human} & {Synthetic} & {OOD} & {Synthetic} & {OOD} & {Synthetic} & {OOD} &\\
    \toprule 
    \multicolumn{5}{l}{\textbf{Classical}} \\
     Avg. CLIP Score ~\cite{clip_radford}  & 0.3  & 0.4 & 0.17 & 0.3 & 0.45 & 0.78 & 0.8 & \\
     Avg. Image Reward Score~\cite{imagereward}  & 0.73  & 0.6 & 0.66 & 0.7 & 0.54 & 0.65 & 0.85 & \\
     Avg. PickScore~\cite{Pick-a-Pic}  & 0.37  & 0.25 & 0.34 & 0.26 & 0.21 & 0.81 & 0.85 & \\
     % Avg. CLIP 3DQ Score & TBD & TBD & TBD & TBD & TBD & \\
    \midrule 
    \multicolumn{5}{l}{\textbf{Vision Large Language models}} \\
     Phi-3.5-Vision~\cite{phi}  & 0.53 & 0.47 & 0.54 & 0.49 & 0.5 & 0.64 & 0.65 & \\
     LLaVA-Qwen-7B~\cite{llava}  & 0.54 & 0.46  & 0.54 & 0.51 & 0.46 & 0.68 & 0.58 &\\
     LLaVA-Llama3-8b~\cite{llava}  & 0.5 & 0.5 & 0.47 & 0.47 &  0.48 & 0.49 & 0.5 & \\
     Llama3.2-Vision-11B*~\cite{llama3}  & 0.06 & 0.04 & 0.05 & 0.1 & 0.07 & 0.04 & 0.5 & \\
     BLIP*~\cite{blip}  & 0.05 & 0.28 & 0.2 & 0.07 & 0.09 & 0.37 & 0.13 &\\
     GPT-4o*~\cite{GPT4}  & 0.59 & 0.48 & 0.69 & 0.54 & 0.54 & 0.61 & 0.55 &\\
     PaliGemma*~\cite{paligemma}  & 0.02 & 0.02 & 0.21 & 0.25 & 0.25 & 0.17 & 0.1 & \\
     \midrule
     \shortmethod (CLIP)  & \textbf{0.9}  & 0.85 & \textbf{0.89} & \textbf{0.99} & \textbf{0.67} & 0.98 & \textbf{0.86} & \\
     \shortmethod (CLIP + Fit3D)  & \textbf{0.9}  & 0.88 & 0.78 & 0.97 & 0.57 & \textbf{1} & 0.53 & \\
     \shortmethod (CLIP + DinoV2)  & 0.86 & \textbf{0.89} & 0.78 & \textbf{0.99} & 0.51 & 0.98 & 0.74 & \\
     w/ Fit3D  & 0.81 & 0.8 & 0.55 & 0.89 & 0.44  & 0.93 & 0.44 &\\
     w/ DinoV2  & 0.77 & 0.87 & 0.54 & 0.97 & 0.61 & 0.75 & 0.58 &\\
    \bottomrule
    \end{tabular}}
\vspace{-0.15cm}
\caption{We report accuracy for curated synthetic and out-of-domain human preference evaluation datasets for appearance, surface and fidelity to text. Additionally, for appearance, we compare methods on in-domain (unseen prompts from methods used for training data). We compare our method against classical metric methods as well as other vLLMs. For text fidelity, we do not provide prompts or pass empty strings for the classical methods. Methods with * next to their names do not currently support multi-image input and were passed either 4x2 grids composed of eight images or 8 input images in sequence in case of GPT-4o. \label{tbl:method_comparison}}
\end{table*}

%% file: tables/leaderboard_methods.tex
% \begin{table}
% \begin{centering}
% \resizebox{\linewidth}{!}{
% \begin{tabular}{lccccc}
%     \multicolumn{1}{l}{\textbf{Methods}} &\multicolumn{1}{c}{Appear.} &\multicolumn{1}{c}{Surf.} & \multicolumn{1}{c}{T-Fidelity} & \multicolumn{1}{c}{Overall} \\
%     \midrule
%      Magic3d~\cite{magic3d} & \cellcolor{color1} 1 & \cellcolor{color1} 1 & \cellcolor{color1} 1  & \cellcolor{color1} 1\\
%      Latentnerf~\cite{latentnerf} & \cellcolor{color2} 2 &  N/A & \cellcolor{color2} 2 & \cellcolor{color2} 2\\
%      Magic123~\cite{Magic123} & \cellcolor{color4} 4 & \cellcolor{color3} 3 & \cellcolor{color6} 6 & \cellcolor{color5} 5 \\
%      Dreamfusion~\cite{dreamfusion} & \cellcolor{color3} 3 & \cellcolor{color3} 2 &\cellcolor{color3} 3 & \cellcolor{color3} 3 \\
%      Vfusion-3d~\cite{vfusion3d} & \cellcolor{color5} 5 &  N/A & \cellcolor{color4} 4 & \cellcolor{color4} 4 \\
%      Flex3d~\cite{flex3d} & \cellcolor{color6} 6 &  N/A & \cellcolor{color5} 5 & \cellcolor{color6} 6\\
%      \bottomrule 
%     \end{tabular}}
%     \end{centering}

\begin{table}
\begin{centering}
\resizebox{0.9\linewidth}{!}{
\begin{tabular}{lccccc}
    \multicolumn{1}{l}{\textbf{Methods}} &\multicolumn{1}{c}{Appear.} &\multicolumn{1}{c}{Surf.} & \multicolumn{1}{c}{T-Fidelity} & \multicolumn{1}{c}{Overall} \\
    \midrule
    Trellis*~\cite{trellis}  & \cellcolor{color1} 1 & \cellcolor{color1} 1 & \cellcolor{color1} 1  &  \cellcolor{color1} 1 \\
    AssetGen~\cite{assetgen}  & \cellcolor{color2} 2 & \cellcolor{color7} 7 & \cellcolor{color2} 2  &  \cellcolor{color2} 2 \\
    Flex3d*~\cite{flex3d} & \cellcolor{color4} 4 &  N/A & \cellcolor{color3} 3 & \cellcolor{color3} 3\\
     Latentnerf~\cite{latentnerf} & \cellcolor{color3} 3 &  N/A & \cellcolor{color6} 6 &  \cellcolor{color4} 4\\
     Magic123~\cite{Magic123} & \cellcolor{color5} 5 & \cellcolor{color4} 4 & \cellcolor{color4} 4 &  \cellcolor{color4} 5 \\
     Vfusion-3d*~\cite{vfusion3d} & \cellcolor{color6} 6 &  \cellcolor{color8} 8 & \cellcolor{color5} 5 &  \cellcolor{color5} 6 \\
     Magic3d~\cite{magic3d} & \cellcolor{color7} 7 & \cellcolor{color5} 5 & \cellcolor{color7} 7  &  \cellcolor{color7} 7 \\
     Dreamfusion~\cite{dreamfusion} & \cellcolor{color8} 8 & \cellcolor{color2} 2 &\cellcolor{color8} 8 & \cellcolor{color8} 8 \\
     Fantasia3D~\cite{fantasia3d}  &  N/A & \cellcolor{color3} 3 &  N/A  &  N/A\\
    TripoSR~\cite{triposr}  &  N/A & \cellcolor{color6} 6 &  N/A  &  N/A\\
     \bottomrule 
    \end{tabular}}
    \end{centering}

\vspace{-0.1cm}
\caption{\shortmethod applied to 3D generation methods on \shortmethod-Bench. Methods are ranked (\colorbox{color1}{Best},\colorbox{color6}{Worst}) on text fidelity, appearance and surface quality score (if available). Only appearance and text fidelity are used for the overall score. Image-to-3D methods are denoted with *.
 \label{tbl:leaderboard_methods}}
\end{table}

% \begin{table}
% \begin{centering}

% \begin{tabular}{lcccccc}
%     &\multicolumn{2}{c}{Appearance} & \multicolumn{1}{c}{Surface} & \multicolumn{2}{c}{T-Fidelity}\\
%     \cmidrule(lr){2-3} \cmidrule(lr){4-4} \cmidrule(lr){5-6}  
%     & {Human} & {Synthetic} & {Synthetic} & {Human} & {Synthetic} & \\
    
%     \midrule
%      \cmidrule(lr){2-3} \cmidrule(lr){4-5} \cmidrule(lr){6-7} 
%     \multicolumn{5}{l}{\textbf{Methods}} \\
%      Magic3d~\cite{magic3d} & \textbf{1558.37} & \textbf{1558.37}  & \\
%      Latentnerf~\cite{metzer22latent-nerf} & 1516.51 & 1500.51  & \\
%      Magic123 (coarse)~\cite{Magic123} & 1515.93 & 1515.93  & \\
%      Dreamfusion~\cite{dreamfusion} & 1470.79 & 1486.79  & \\
%      Vfusion-3d~\cite{vfusion3d} & 1438.39 & 1438.39 & \\
%      \bottomrule 
%     \end{tabular}
%     \end{centering}

% \caption{Ranking output metric for text faithfulness and 3D-Quality of \shortmethod applied to 5 3D Evaluation methods on \shortmethod-BenchS prompts.
%  \label{tbl:leaderboard_methods}}
 
% \end{table}

%% file: sec/8_conclusion.tex
% Conclusion:
% We did it. This paper rocks and you are lucky to have read it (i.e. brief recap of the entire paper). Also, we’ll do all these other amazing things in the future. 
% To keep going with the analogy, you can think of future work as (potential) academic offspring (credits to James).

\section{Conclusion}
In this paper, we have laid out the lack of an existing metric in the text-to-3D domain that caters to all its necessary parameters and established the relevance for developing such a metric. It is a difficult problem to solve because of the many ways in which 3D generation is supported. Diverging from the trend of using similarity metrics, which is impractical in the case of text-to-3D generation, we propose introducing 3D aesthetic preference into the vision-language space and transforming that into a ready-to-use evaluation ranking metric using vLLMs. We demonstrate that \shortmethod, a vLLM containing 8.35 billion parameters and trained using a synthetically curated 3D-object dataset coupled with user preference data establishes itself as a comprehensive, accessible and competitive evaluation method displaying strong performance on relevant metric dimensions, i.e., appearance, texture and text fidelity, in alignment with user 3D object preference.
We hope that \shortmethod will provide a standard benchmark and metric for the comparison of existing and future methods. 

%% file: sec/X_suppl.tex
\clearpage
\setcounter{page}{1}
\maketitlesupplementary

\section{Training Dataset}
\subsection{Examples from the Pre-training Dataset}
In this section, we present examples from the pre-training dataset utilized to train \shortmethod. Figures \ref{fig:suppl_pre_1} and \ref{fig:suppl_pre_2} illustrate the different types of input data generated from a single 3D asset, accompanied by the corresponding Question-Answer prompts. 

\input{figures/supp_pre_fig1}
\input{figures/supp_pre_fig2}

\subsection{Examples from the Supervised Fine-tuning Dataset}
Figures \ref{fig:suppl_sft_1} and \ref{fig:suppl_sft_2} provide more examples from the supervised fine-tuning dataset employed in training \shortmethod. The SFT dataset distribution is displayed in Figure~\ref{fig:finetuning_dataset}.

\input{figures/supp_sft_fig1}
\input{figures/supp_sft_fig2}
\input{figures/finetuning_dataset}

\section{Dataset Ablation}
\begin{figure}[H]
        \centering
            \includegraphics[width=\linewidth]{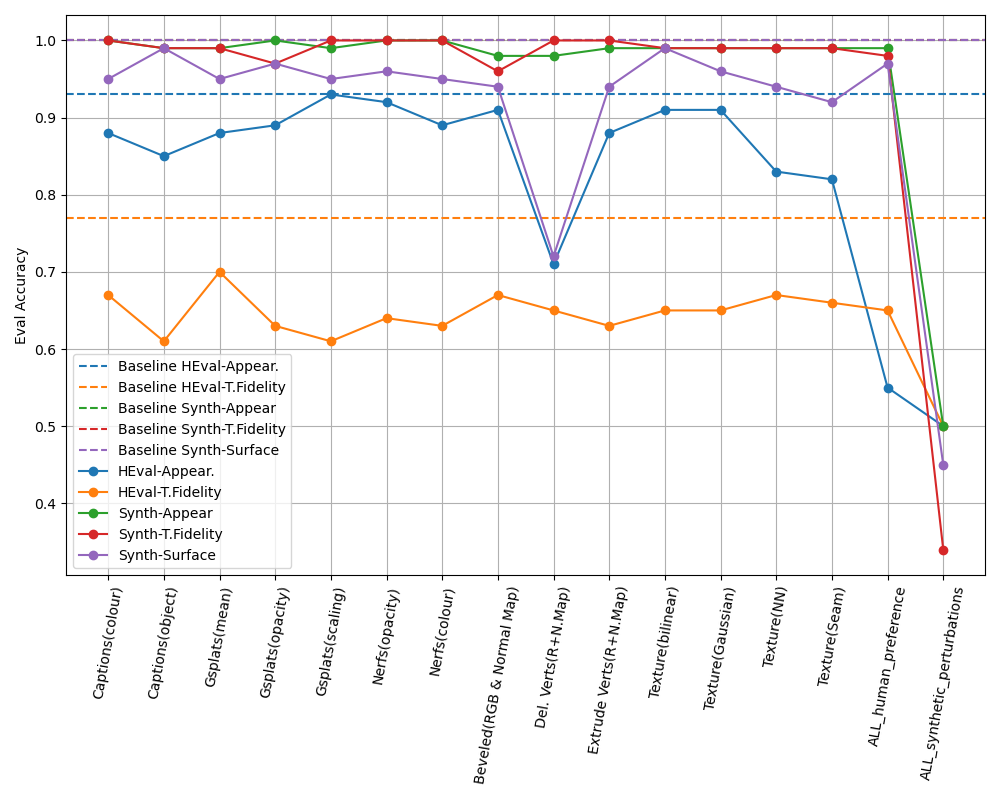}
           \caption{Ablation on training data by REMOVING subsets of data from the final fine-tuning dataset and evaluated on the held-out evaluation datasets. Dotted lines: accuracy when fine-tuned using the entire SFT dataset (same random seed). }
           \label{fig:ablation_rebuttal}
    \end{figure}

Please refer to Figure \ref{fig:ablation_rebuttal} to see how the performance of \shortmethod varies with the removal of different subgroups of the dataset. 

\section{Limitations}
\shortmethod exhibits erratic performance when there is Janus (subparts repeated in an object like multiple faces), and on out-of-domain surface evaluation. Figure \ref{fig:supp_limitations} provides an example of this limitation, where the method compares multiple assets generated from the same prompt and ranks them from best to worst.

\input{figures/supp_limitation_fig1}

\section{Benchmark Analysis}
Please refer to Table \ref{tbl:benchmark_comparison} for comparison of \shortmethod-Bench with different existing benchmarks of generation prompts. Our aim was to come up with a small and diverse dataset containing an even split in terms of object type (animate like humanoids and animals vs. inanimate such as chairs, tables, football) and composition (single vs composite objects or scenes) to allow for granularity in the evaluation of 3D assets. Moreover, we also wanted to increase the mean and variance for the length of the prompts.
\input{tables/benchmark_analysis}

\section{Comparison of objects generated from different prompts}
Since \shortmethod disambiguates evaluation on the basis of 1) Appearance 2) Text Fidelity and 3) Surface Quality, we additionally test its performance on a benchmark containing pairs of 3D objects generated by different prompts and annotating on the basis of appearance and filtered to remove any ambiguous samples. \shortmethod has an accuracy of 0.88 on this benchmark. Qualitative examples are provided in Figure \ref{fig:supp_random_obj} .

\input{figures/supp_random_obj_comp}

\section{Results}
\subsection{Qualitative ablation study for \shortmethod's image encoder choices}
% Check this out later
In our ablation study involving different image encoders, we evaluated the quantitative metrics of using CLIP, DinoV2, and Fit3D~\cite{fit3d}, as well as combinations of these with CLIP~\cite{clip_radford}. Our findings indicate that while the \shortmethod with CLIP consistently performed well across all evaluation datasets, the pairing of DinoV2~\cite{dinov2} and CLIP was not too far behind. On investigating fuurther, we noticed that \shortmethod with CLIP and DinoV2 gave more weight to 3D coherence and plausibility where \shortmethod with standalone CLIP leans towards more visually appealing objects. Given that these image embeddings capture distinct object features, we provide qualitative examples generated on \shortmethod-Bench to compare and contrast \shortmethod's asset preferences. Figure \ref{fig:supp_imenc_comp2} contrasts the strengths and weaknesses of \shortmethod with CLIP and with the combination of CLIP and DinoV2 respectively. Overall, both embeddings capture relevant 3D features for comparison as shown in Figure \ref{fig:supp_imenc_comp1}.

\input{figures/supp_imenc_comp_fig2}
\input{figures/supp_imenc_comp_fig1}

\subsection{Qualitative Comparison of Leaderboard Methods}
We present qualitative examples of pairwise evaluation and ranking of generative 3D methods from our leaderboard on \shortmethod-Bench. These examples focus on their performance on appearance (Figure \ref{fig:suppl_leaderboard_1}), surface quality (Figure \ref{fig:suppl_leaderboard_2}), and text faithfulness (Figure \ref{fig:suppl_leaderboard_3}).

\input{figures/supp_leaderboard_fig1}
\input{figures/supp_leaderboard_fig2}
\input{figures/supp_leaderboard_fig3}

%% file: figures/supp_pre_fig1.tex
\begin{figure}[H]
% \vspace{-2cm}
\begin{center}
\includegraphics[width=\linewidth]{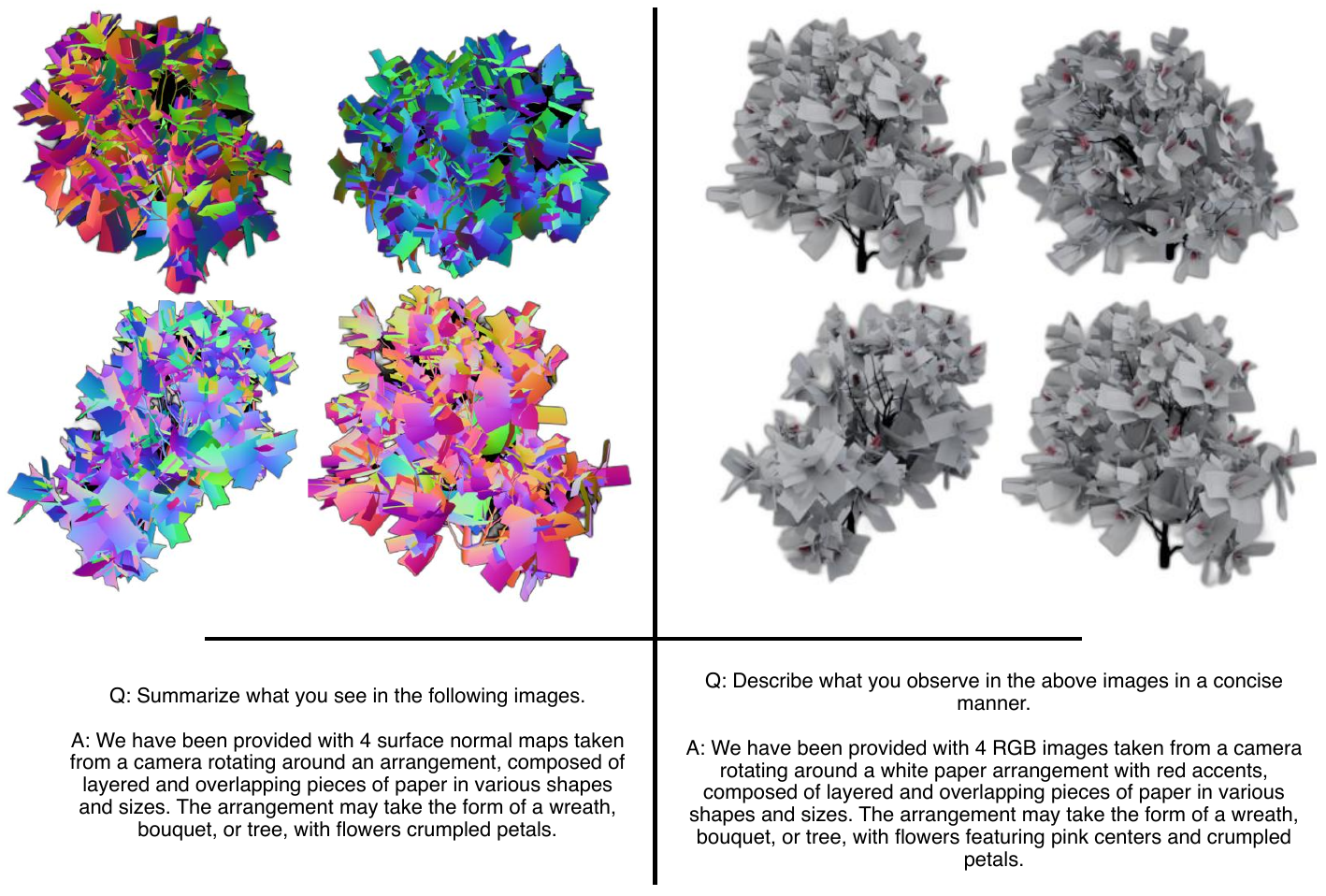}
\end{center}
% \vspace{-0.5cm}
\caption{\textbf{Pre-training Dataset} We use multiple views of RGB and surface normal maps rendered from a 3D object, accompanied by a Question-Answer prompt that summarizes the object. }

\label{fig:suppl_pre_1}
\end{figure}

%% file: figures/supp_pre_fig2.tex
\begin{figure}[H]
% \vspace{-1cm}
\begin{center}
\includegraphics[width=1\linewidth]{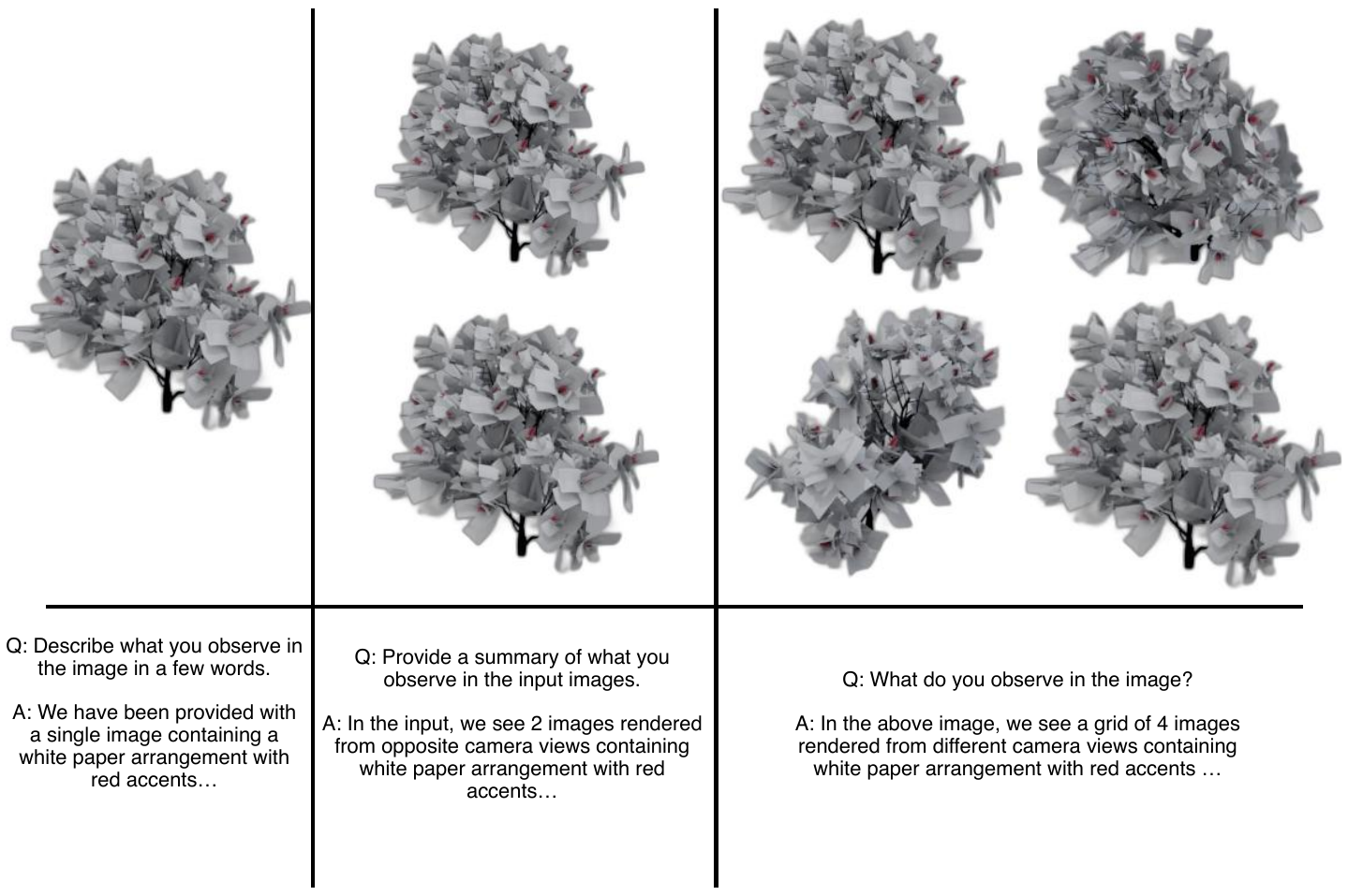}
\end{center}
% \vspace{-0.5cm}
\caption{\textbf{Pre-training Dataset} We use single and multiple views rendered from a 3D object as well as an image grid composed of the aforementioned multi-view (4) RGB images. }

\label{fig:suppl_pre_2}
\end{figure}

%% file: figures/supp_sft_fig1.tex
\begin{figure}[H]
% \vspace{-2.6cm}
\begin{center}
\includegraphics[width=1.\linewidth]{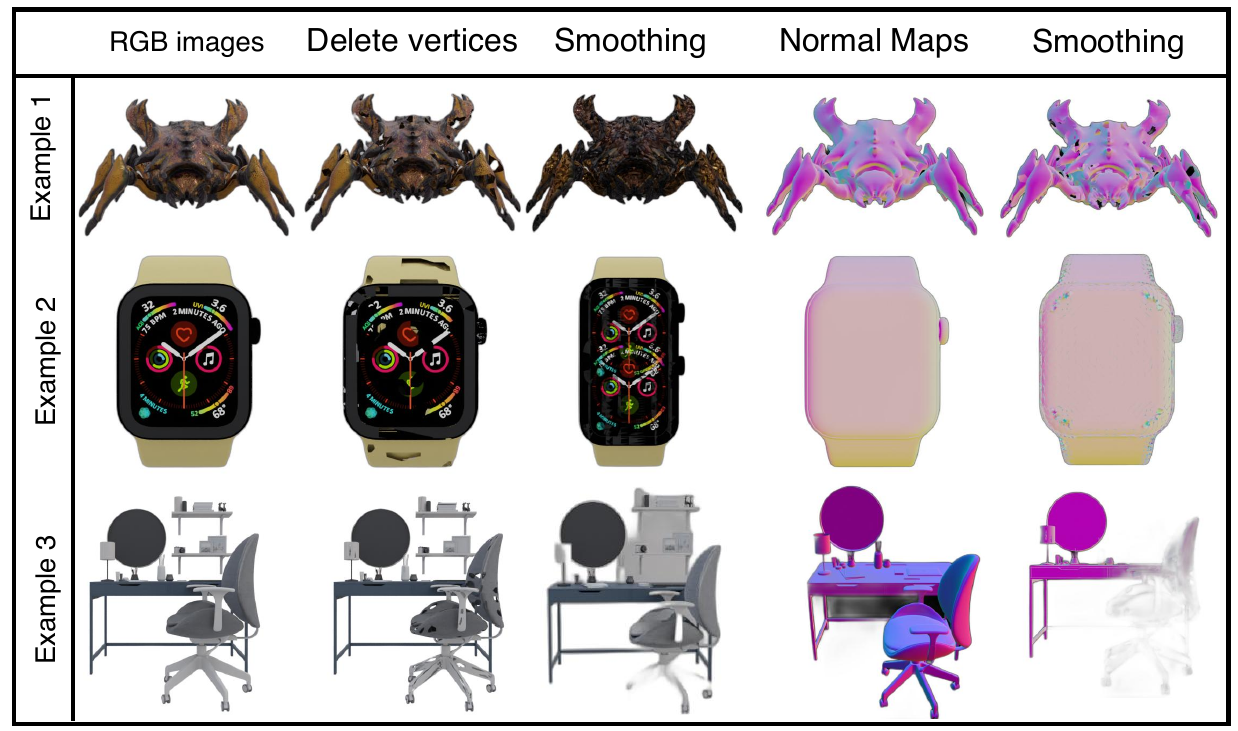}
\end{center}
% \vspace{-0.5cm}
\caption{\textbf{Deep dive into Supervised Fine-tuning Dataset:} We use single and multiple views of RGB and surface normals rendered from a 3D object generated from a prompt. Further, we take these objects and perturb them to simulate common appearance, surface and text-related artefacts in generative 3D methods. In this figure, we showcase Laplacian smoothing and random deletion of vertices in the original meshes.}

\label{fig:suppl_sft_1}
\end{figure}

%% file: figures/supp_sft_fig2.tex
\begin{figure}[H]
% \vspace{-1.cm}
\begin{center}
\includegraphics[width=1.\linewidth]{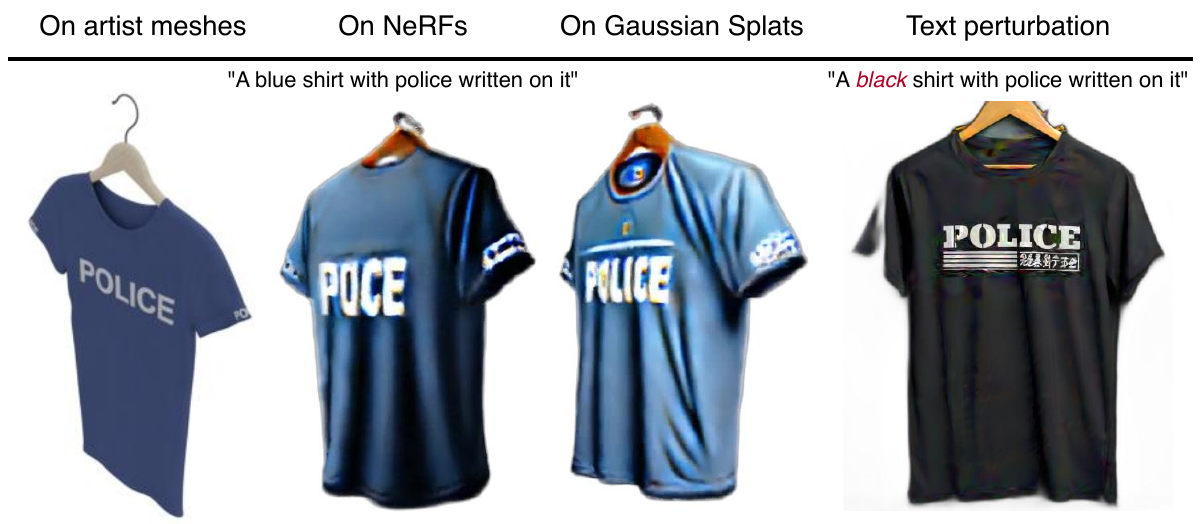}
\end{center}
% \vspace{-0.5cm}
\caption{\textbf{Deep dive into Supervised Fine-tuning Dataset:} Further, we use artist-drawn meshes of 3D objects and perturb them to simulate common appearance, surface and text-related artefacts in generative 3D methods. In this figure, we showcase textual and structure specific perturbations, i.e., by generating objects using NeRFs and Gaussian splatting.}

\label{fig:suppl_sft_2}
\end{figure}

%% file: figures/finetuning_dataset.tex
\begin{figure}
% \vspace{-0.20cm}
\begin{center}
\includegraphics[width=1\linewidth]{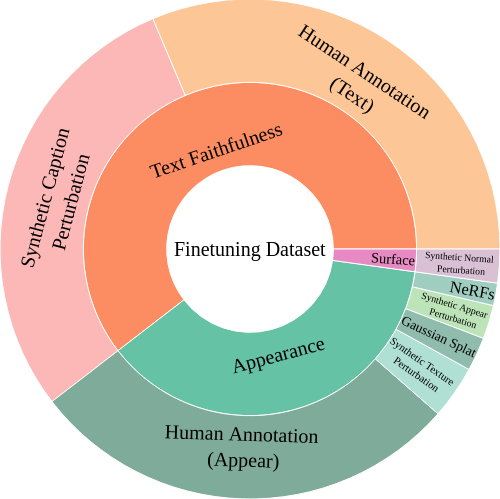}
% \vspace{-0.52cm}
\end{center}
\caption{\textbf{Data distribution} for the SFT dataset used in training \shortmethod. It consists of appearance, surface quality and text fidelity comparison data that are synthetically generated from artist-created meshes as well curated from user annotation with outputs from text-to-3D methods.}
% \vspace{-2mm}
\label{fig:finetuning_dataset}
\end{figure}

%% file: figures/supp_limitation_fig1.tex
\begin{figure}[H]
% \vspace{-4cm}
\begin{center}
\includegraphics[width=1\linewidth]{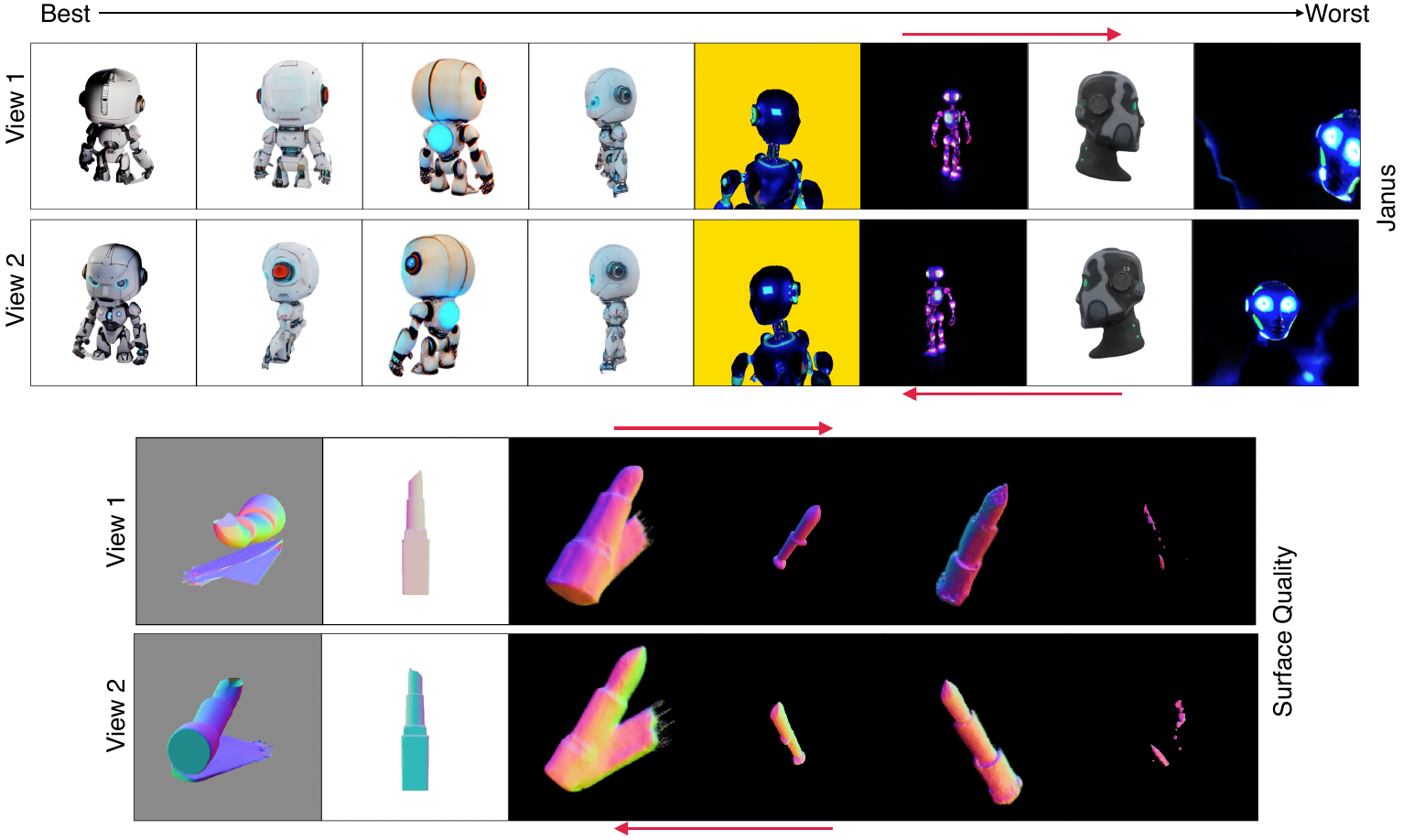}
\end{center}
% \vspace{-0.5cm}
\caption{\textbf{Limitation of \shortmethod.}  \shortmethod has limited success detecting janus and out-of-domain surface normal images. The image shows how \shortmethod ranks the objects. The red arrow points to the expected ranking of the object.}
\label{fig:supp_limitations}
\end{figure}

%% file: tables/benchmark_analysis.tex
\begin{table*}[!hbt]
\centering
\resizebox{1\linewidth}{!}{\begin{tabular}{lccccccc}
    &\multicolumn{2}{c}{General} & \multicolumn{2}{c}{Object Type} & \multicolumn{2}{c}{Composition}\\
    \cmidrule(lr){2-3} \cmidrule(lr){4-5} \cmidrule(lr){6-7}  
    & {Num. Prompts} & {Avg. word length} & {Animate} & {Inanimate} & {Single Obj} & {Multi-object} & \\
    \toprule 
    T3Bench~\cite{t3bench} & 300  & 7.98 & 36 & 264 & 100 & 200 &  \\
    ChatGPTEval3D~\cite{Gpt4vEval} & 110  & 11.49 & 18 & 92 & 65 & 45 &  \\
    DreamFusion~\cite{dreamfusion} & 404  & 6.98 & 211 & 192 & 154 & 250 &  \\
    \textbf{\shortmethod-Bench} & 80  & 12.863 & 40 & 40 & 43 & 37 &  \\
    \bottomrule
    \end{tabular}}
\caption{\textbf{Comparing \shortmethod-Bench with existing 3D generation prompt benchmarks}. \label{tbl:benchmark_comparison}}
\end{table*}

%% file: figures/supp_random_obj_comp.tex
\begin{figure}[H]
% \vspace{-1cm}
\begin{center}
\includegraphics[width=1.\linewidth]{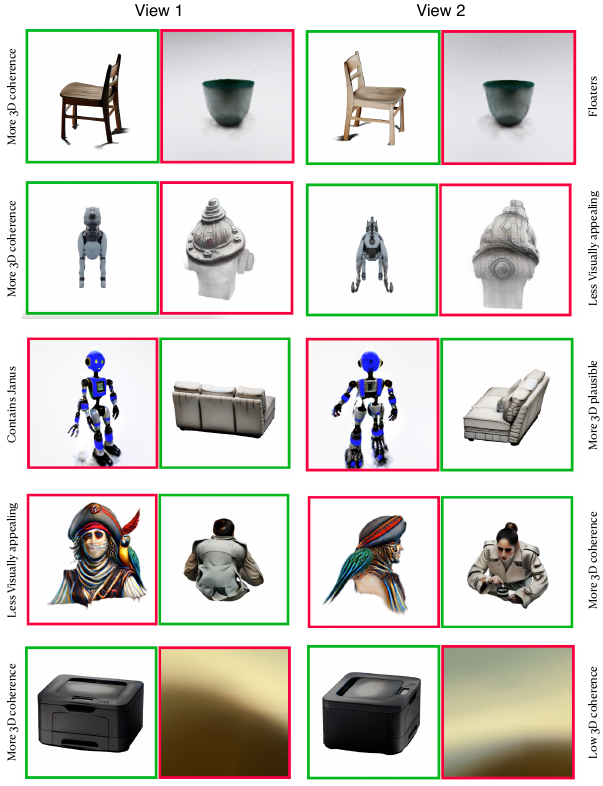}
\end{center}
% \vspace{-0.5cm}
\caption{\textbf{Qualitative result from \shortmethod on comparing objects generated from different text prompts on Appearance.} This image displays 5 examples of the preference of \shortmethod from an annotated evaluation dataset where we conduct pairwise comparison of objects generated from different text prompts on the basis of their appearance only. Green border is for the preferred object and red for the other object. We use 4 views as input but in the image, we display two views side-by-side.}

\label{fig:supp_random_obj}
\end{figure}

%% file: figures/supp_imenc_comp_fig2.tex
\begin{figure}[H]
% \vspace{-1.cm}
\begin{center}
\includegraphics[width=1\linewidth]{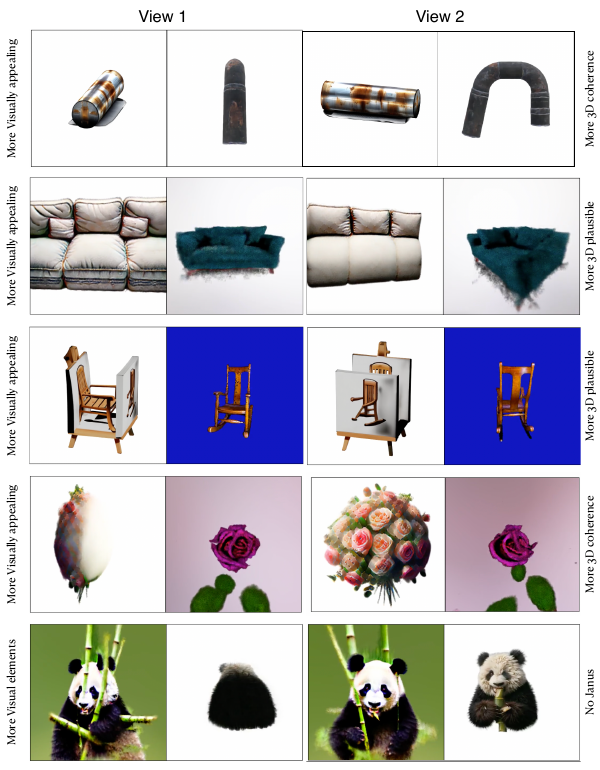}
\end{center}
% \vspace{-0.5cm}
\caption{\textbf{Qualitative result from ablation study on different input image embeddings on Appearance.} 
Object Preference - \textit{L.H.S} (\shortmethod w/ CLIP), \textit{R.H.S} (\shortmethod w/ CLIP+DinoV2): We demonstrate five examples containing (displaying two views side-by-side to provide some clarity). Observation: CLIP evaluates more favourably on visual/appearance/surface properties whereas CLIP+DinoV2 prefers more on the basis of 3D coherency (lack of janus) and plausibility.
}

\label{fig:supp_imenc_comp2}
\end{figure}

%% file: figures/supp_imenc_comp_fig1.tex
\begin{figure*}
% \vspace{-1cm}
\begin{center}
\includegraphics[width=0.7\linewidth]{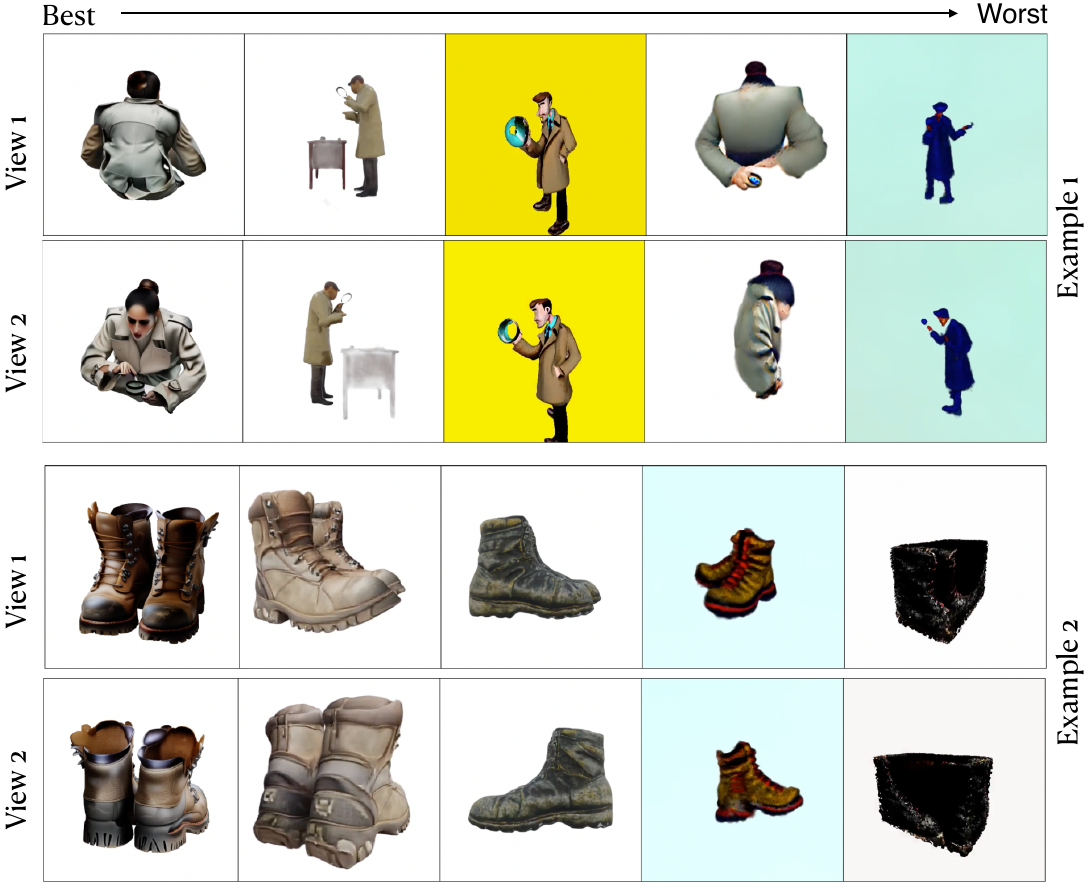}
\end{center}
% \vspace{-0.5cm}
\caption{\textbf{Qualitative result from ablation study on different input image embeddings on Appearance.} This image displays a degree of correlation between the preferences of \shortmethod when using either standalone CLIP embeddings or CLIP combined with DinoV2 for two examples containing (displaying two views to provide some clarity), since both the encoders select these assets in the same order using two examples where the objects were ranked in a similar manner. }

\label{fig:supp_imenc_comp1}
\end{figure*}

%% file: figures/supp_leaderboard_fig1.tex
\begin{figure*}[!hbt]
% \vspace{-1.cm}
\begin{center}
\includegraphics[width=0.7\linewidth]{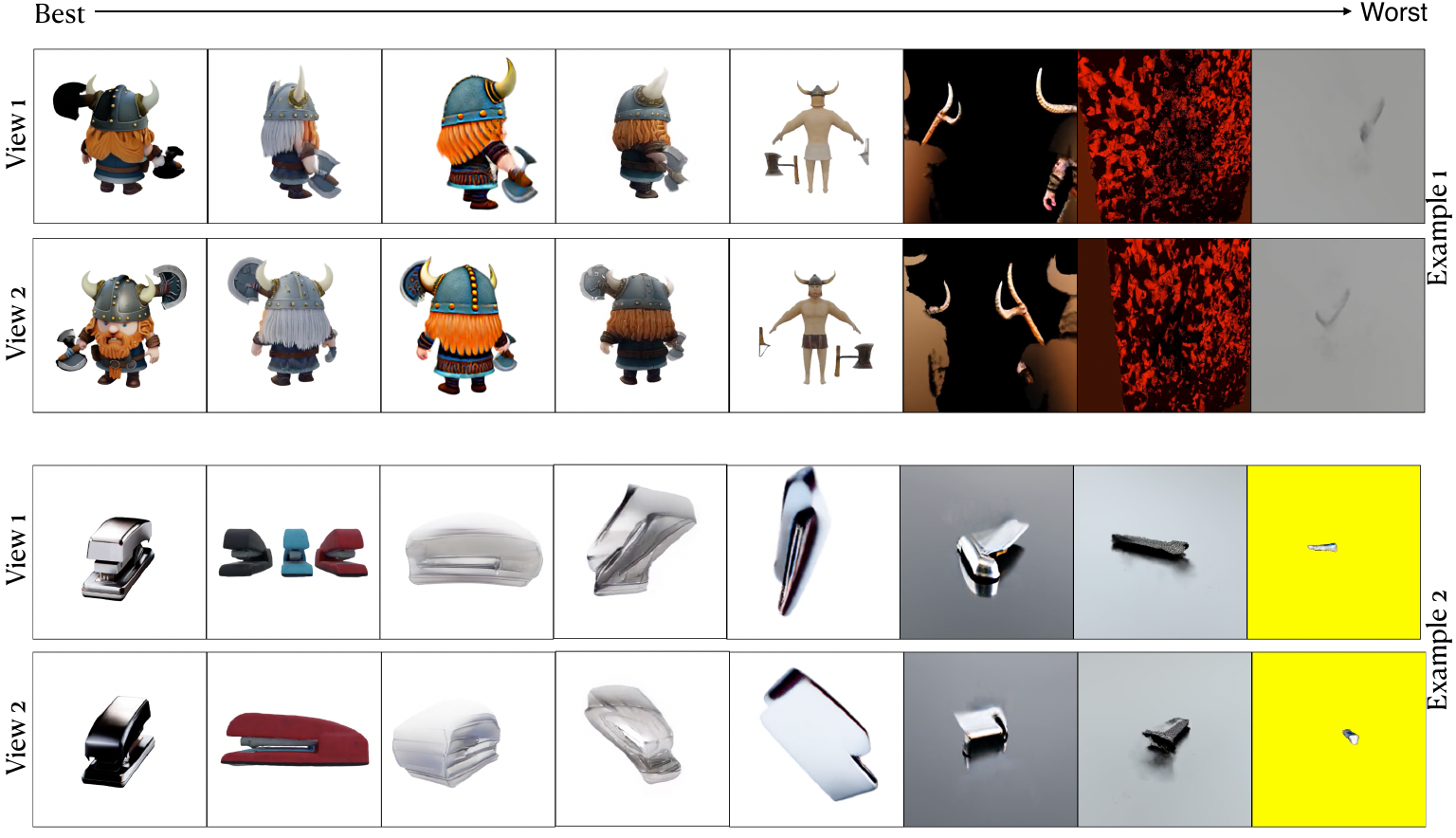}
\end{center}
% \vspace{-0.5cm}
\caption{\textbf{Comparison of leaderboard methods (Appearance).} Qualitative examples from applying \shortmethod to evaluate 3D generative methods on appearance quality parameter. Left to Right: Best object to worst in pairwise comparison of all assets for the same prompt.}

\label{fig:suppl_leaderboard_1}
\end{figure*}

%% file: figures/supp_leaderboard_fig2.tex
\begin{figure*}[!hbt]
% \vspace{.5cm}
\begin{center}
\includegraphics[width=0.7\linewidth]{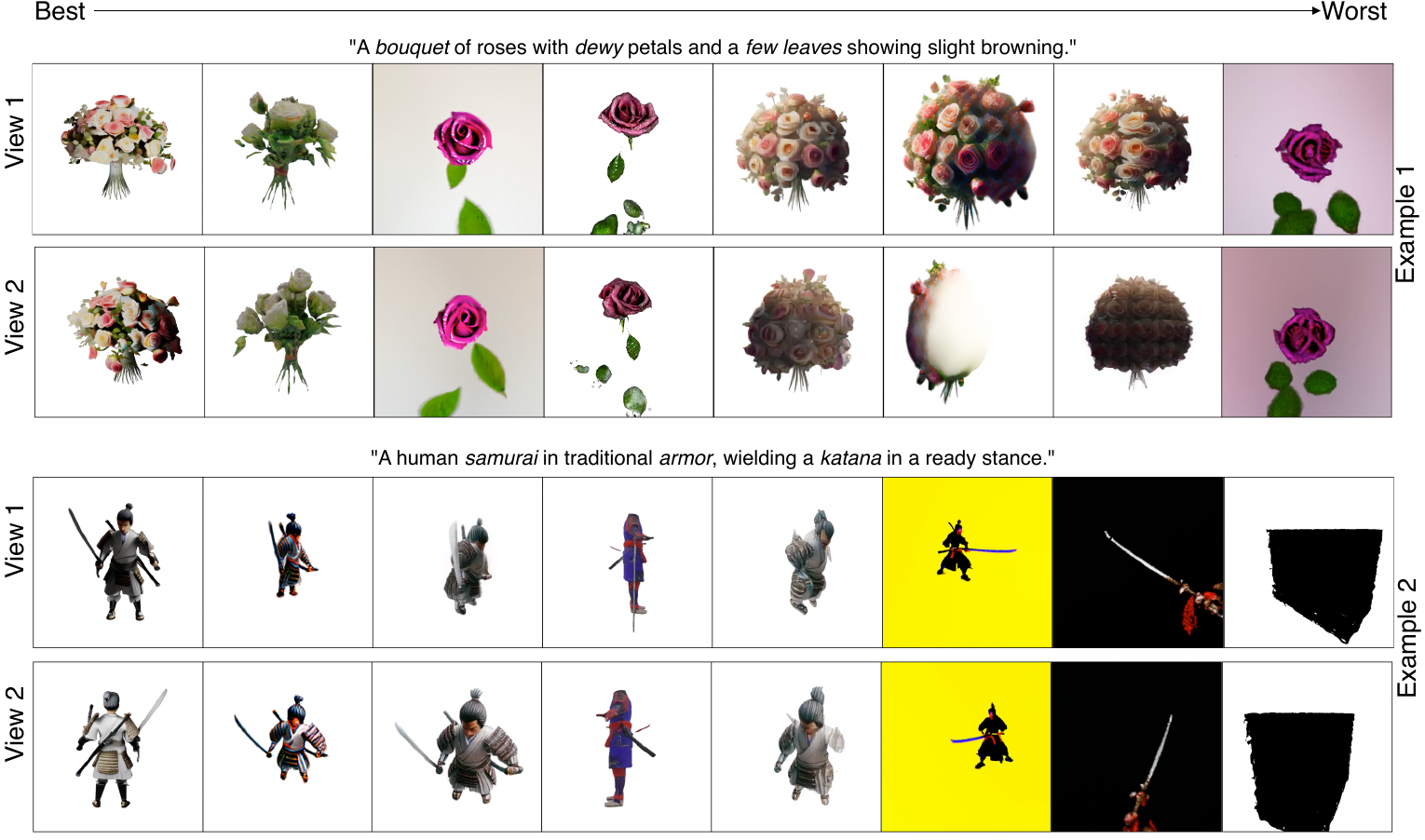}
\end{center}
% \vspace{-0.5cm}
\caption{\textbf{Comparison of leaderboard methods (Text Fidelity).} Qualitative examples from applying \shortmethod to evaluate 3D generative methods on the text fidelity parameter. Left to Right: Best object to worst in pairwise comparison of all assets for the same prompt. }

\label{fig:suppl_leaderboard_2}
\end{figure*}

%% file: figures/supp_leaderboard_fig3.tex
\begin{figure}[H]
% \vspace{-2.cm}
\begin{center}
\includegraphics[width=1\linewidth]{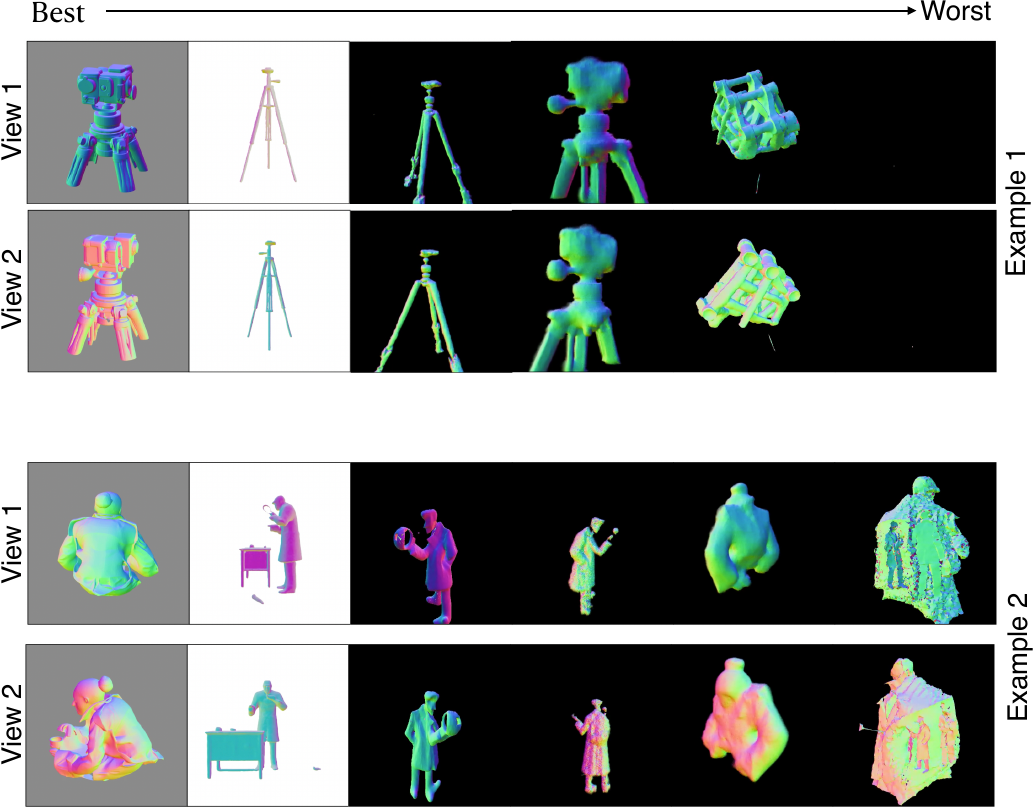}
\end{center}
% \vspace{-0.5cm}
\caption{\textbf{Comparison of leaderboard methods (Surface Quality).} Qualitative examples from applying \shortmethod to evaluate 3D generative methods on the surface quality parameter. Left to Right: Best object to worst in pairwise comparison of all assets for the same prompt.}

\label{fig:suppl_leaderboard_3}
\end{figure}

%% file: main.bbl
\begin{thebibliography}{64}
\providecommand{\natexlab}[1]{#1}
\providecommand{\url}[1]{\texttt{#1}}
\expandafter\ifx\csname urlstyle\endcsname\relax
  \providecommand{\doi}[1]{doi: #1}\else
  \providecommand{\doi}{doi: \begingroup \urlstyle{rm}\Url}\fi

\bibitem[Abdin et~al.(2024)Abdin, Aneja, Awadalla, Awadallah, Awan, Bach, Bahree, Bakhtiari, Bao, Behl, et~al.]{phi}
Marah Abdin, Jyoti Aneja, Hany Awadalla, Ahmed Awadallah, Ammar~Ahmad Awan, Nguyen Bach, Amit Bahree, Arash Bakhtiari, Jianmin Bao, Harkirat Behl, et~al.
\newblock Phi-3 technical report: A highly capable language model locally on your phone.
\newblock \emph{arXiv preprint arXiv:2404.14219}, 2024.

\bibitem[AI(2024)]{llama3}
Meta AI.
\newblock The llama 3 herd of models, 2024.

\bibitem[Anthropic()]{Claude}
Anthropic.

\bibitem[Bai et~al.(2023)Bai, Bai, Chu, Cui, Dang, Deng, Fan, Ge, Han, Huang, Hui, Ji, Li, Lin, Lin, Liu, Liu, Lu, Lu, Ma, Men, Ren, Ren, Tan, Tan, Tu, Wang, Wang, Wang, Wu, Xu, Xu, Yang, Yang, Yang, Yang, Yao, Yu, Yuan, Yuan, Zhang, Zhang, Zhang, Zhang, Zhou, Zhou, Zhou, and Zhu]{qwen}
Jinze Bai, Shuai Bai, Yunfei Chu, Zeyu Cui, Kai Dang, Xiaodong Deng, Yang Fan, Wenbin Ge, Yu Han, Fei Huang, Binyuan Hui, Luo Ji, Mei Li, Junyang Lin, Runji Lin, Dayiheng Liu, Gao Liu, Chengqiang Lu, Keming Lu, Jianxin Ma, Rui Men, Xingzhang Ren, Xuancheng Ren, Chuanqi Tan, Sinan Tan, Jianhong Tu, Peng Wang, Shijie Wang, Wei Wang, Shengguang Wu, Benfeng Xu, Jin Xu, An Yang, Hao Yang, Jian Yang, Shusheng Yang, Yang Yao, Bowen Yu, Hongyi Yuan, Zheng Yuan, Jianwei Zhang, Xingxuan Zhang, Yichang Zhang, Zhenru Zhang, Chang Zhou, Jingren Zhou, Xiaohuan Zhou, and Tianhang Zhu.
\newblock Qwen technical report.
\newblock \emph{arXiv preprint arXiv:2309.16609}, 2023.

\bibitem[Bavishi et~al.(2023)Bavishi, Elsen, Hawthorne, Nye, Odena, Somani, and Ta\c{s}\i{}rlar]{fuyu}
Rohan Bavishi, Erich Elsen, Curtis Hawthorne, Maxwell Nye, Augustus Odena, Arushi Somani, and Sa\u{g}nak Ta\c{s}\i{}rlar.
\newblock Introducing our multimodal models, 2023.

\bibitem[Bensadoun et~al.(2024)Bensadoun, Kleiman, Azuri, Harosh, Vedaldi, Neverova, and Gafni]{texturegen}
Raphael Bensadoun, Yanir Kleiman, Idan Azuri, Omri Harosh, Andrea Vedaldi, Natalia Neverova, and Oran Gafni.
\newblock Meta 3d texturegen: Fast and consistent texture generation for 3d objects.
\newblock \emph{arXiv preprint arXiv:2407.02430}, 2024.

\bibitem[Beyer et~al.(2024)Beyer, Steiner, Pinto, Kolesnikov, Wang, Salz, Neumann, Alabdulmohsin, Tschannen, Bugliarello, et~al.]{paligemma}
Lucas Beyer, Andreas Steiner, Andr{\'e}~Susano Pinto, Alexander Kolesnikov, Xiao Wang, Daniel Salz, Maxim Neumann, Ibrahim Alabdulmohsin, Michael Tschannen, Emanuele Bugliarello, et~al.
\newblock Paligemma: A versatile 3b vlm for transfer.
\newblock \emph{arXiv preprint arXiv:2407.07726}, 2024.

\bibitem[Bruce et~al.(2024)Bruce, Dennis, Edwards, Parker-Holder, Shi, Hughes, Lai, Mavalankar, Steigerwald, Apps, et~al.]{genie}
Jake Bruce, Michael~D Dennis, Ashley Edwards, Jack Parker-Holder, Yuge Shi, Edward Hughes, Matthew Lai, Aditi Mavalankar, Richie Steigerwald, Chris Apps, et~al.
\newblock Genie: Generative interactive environments.
\newblock In \emph{Forty-first International Conference on Machine Learning}, 2024.

\bibitem[Chen et~al.(2023)Chen, Chen, Jiao, and Jia]{fantasia3d}
Rui Chen, Yongwei Chen, Ningxin Jiao, and Kui Jia.
\newblock Fantasia3d: Disentangling geometry and appearance for high-quality text-to-3d content creation.
\newblock In \emph{Proceedings of the IEEE/CVF International Conference on Computer Vision (ICCV)}, pages 22246--22256, 2023.

\bibitem[Deitke et~al.(2023{\natexlab{a}})Deitke, Liu, Wallingford, Ngo, Michel, Kusupati, Fan, Laforte, Voleti, Gadre, VanderBilt, Kembhavi, Vondrick, Gkioxari, Ehsani, Schmidt, and Farhadi]{objaverseXL}
Matt Deitke, Ruoshi Liu, Matthew Wallingford, Huong Ngo, Oscar Michel, Aditya Kusupati, Alan Fan, Christian Laforte, Vikram Voleti, Samir~Yitzhak Gadre, Eli VanderBilt, Aniruddha Kembhavi, Carl Vondrick, Georgia Gkioxari, Kiana Ehsani, Ludwig Schmidt, and Ali Farhadi.
\newblock Objaverse-xl: A universe of 10m+ 3d objects.
\newblock \emph{arXiv preprint arXiv:2307.05663}, 2023{\natexlab{a}}.

\bibitem[Deitke et~al.(2023{\natexlab{b}})Deitke, Schwenk, Salvador, Weihs, Michel, VanderBilt, Schmidt, Ehsani, Kembhavi, and Farhadi]{objaverse}
Matt Deitke, Dustin Schwenk, Jordi Salvador, Luca Weihs, Oscar Michel, Eli VanderBilt, Ludwig Schmidt, Kiana Ehsani, Aniruddha Kembhavi, and Ali Farhadi.
\newblock Objaverse: A universe of annotated 3d objects.
\newblock In \emph{Proceedings of the IEEE/CVF Conference on Computer Vision and Pattern Recognition}, pages 13142--13153, 2023{\natexlab{b}}.

\bibitem[Dosovitskiy et~al.(2020)Dosovitskiy, Beyer, Kolesnikov, Weissenborn, Zhai, Unterthiner, Dehghani, Minderer, Heigold, Gelly, et~al.]{vit}
Alexey Dosovitskiy, Lucas Beyer, Alexander Kolesnikov, Dirk Weissenborn, Xiaohua Zhai, Thomas Unterthiner, Mostafa Dehghani, Matthias Minderer, Georg Heigold, Sylvain Gelly, et~al.
\newblock An image is worth 16x16 words.
\newblock \emph{arXiv preprint arXiv:2010.11929}, 7, 2020.

\bibitem[Elo and Sloan(1978)]{elo_rating}
Arpad~E Elo and Sam Sloan.
\newblock The rating of chessplayers: Past and present.
\newblock \emph{(No Title)}, 1978.

\bibitem[Gao* et~al.(2024)Gao*, Holynski*, Henzler, Brussee, Martin-Brualla, Srinivasan, Barron, and Poole*]{cat3d}
Ruiqi Gao*, Aleksander Holynski*, Philipp Henzler, Arthur Brussee, Ricardo Martin-Brualla, Pratul~P. Srinivasan, Jonathan~T. Barron, and Ben Poole*.
\newblock Cat3d: Create anything in 3d with multi-view diffusion models.
\newblock \emph{Advances in Neural Information Processing Systems}, 2024.

\bibitem[Han et~al.(2024{\natexlab{a}})Han, Kokkinos, and Torr]{vfusion3d}
Junlin Han, Filippos Kokkinos, and Philip Torr.
\newblock Vfusion3d: Learning scalable 3d generative models from video diffusion models, 2024{\natexlab{a}}.

\bibitem[Han et~al.(2024{\natexlab{b}})Han, Wang, Vedaldi, Torr, and Kokkinos]{flex3d}
Junlin Han, Jianyuan Wang, Andrea Vedaldi, Philip Torr, and Filippos Kokkinos.
\newblock Flex3d: Feed-forward 3d generation with flexible reconstruction model and input view curation.
\newblock \emph{arXiv preprint arXiv:2410.00890}, 2024{\natexlab{b}}.

\bibitem[He et~al.(2023)He, Bai, Lin, Zhao, Hu, Sheng, Yi, Li, and Liu]{t3bench}
Yuze He, Yushi Bai, Matthieu Lin, Wang Zhao, Yubin Hu, Jenny Sheng, Ran Yi, Juanzi Li, and Yong-Jin Liu.
\newblock T$^3$bench: Benchmarking current progress in text-to-3d generation, 2023.

\bibitem[He and Wang(2023)]{openlrm}
Zexin He and Tengfei Wang.
\newblock Openlrm: Open-source large reconstruction models.
\newblock \url{https://github.com/3DTopia/OpenLRM}, 2023.

\bibitem[Herrmann(1976)]{laplacian_smoothing}
Leonard~R Herrmann.
\newblock Laplacian-isoparametric grid generation scheme.
\newblock \emph{Journal of the Engineering Mechanics Division}, 102\penalty0 (5):\penalty0 749--756, 1976.

\bibitem[Heusel et~al.(2017)Heusel, Ramsauer, Unterthiner, Nessler, and Hochreiter]{fid}
Martin Heusel, Hubert Ramsauer, Thomas Unterthiner, Bernhard Nessler, and Sepp Hochreiter.
\newblock Gans trained by a two time-scale update rule converge to a local nash equilibrium.
\newblock \emph{Advances in neural information processing systems}, 30, 2017.

\bibitem[H{\"o}llein et~al.(2024)H{\"o}llein, Bo\v{z}i\v{c}, M{\"u}ller, Novotny, Tseng, Richardt, Zollh{\"o}fer, and Nie{\ss}ner]{viewdiff}
Lukas H{\"o}llein, Alja\v{z} Bo\v{z}i\v{c}, Norman M{\"u}ller, David Novotny, Hung-Yu Tseng, Christian Richardt, Michael Zollh{\"o}fer, and Matthias Nie{\ss}ner.
\newblock Viewdiff: 3d-consistent image generation with text-to-image models.
\newblock In \emph{Proceedings of the IEEE/CVF Conference on Computer Vision and Pattern Recognition}, 2024.

\bibitem[Hong et~al.(2023)Hong, Zhang, Gu, Bi, Zhou, Liu, Liu, Sunkavalli, Bui, and Tan]{lrm}
Yicong Hong, Kai Zhang, Jiuxiang Gu, Sai Bi, Yang Zhou, Difan Liu, Feng Liu, Kalyan Sunkavalli, Trung Bui, and Hao Tan.
\newblock Lrm: Large reconstruction model for single image to 3d.
\newblock \emph{arXiv preprint arXiv:2311.04400}, 2023.

\bibitem[Huang et~al.(2023)Huang, Wang, Shi, Tang, Qi, and Zhang]{DreamTimeAI}
Yukun Huang, Jianan Wang, Yukai Shi, Boshi Tang, Xianbiao Qi, and Lei Zhang.
\newblock Dreamtime: An improved optimization strategy for diffusion-guided 3d generation.
\newblock \emph{arXiv preprint arXiv:2306.12422}, 2023.

\bibitem[Jang and Agapito(2024)]{nvist}
Wonbong Jang and Lourdes Agapito.
\newblock Nvist: In the wild new view synthesis from a single image with transformers.
\newblock In \emph{Proceedings of the IEEE/CVF Conference on Computer Vision and Pattern Recognition (CVPR)}, pages 10181--10193, 2024.

\bibitem[Kerbl et~al.(2023)Kerbl, Kopanas, Leimk{\"u}hler, and Drettakis]{kerbl3Dgaussians}
Bernhard Kerbl, Georgios Kopanas, Thomas Leimk{\"u}hler, and George Drettakis.
\newblock 3d gaussian splatting for real-time radiance field rendering.
\newblock \emph{ACM Transactions on Graphics}, 42\penalty0 (4), 2023.

\bibitem[Kirstain et~al.(2023)Kirstain, Polyak, Singer, Matiana, Penna, and Levy]{Pick-a-Pic}
Yuval Kirstain, Adam Polyak, Uriel Singer, Shahbuland Matiana, Joe Penna, and Omer Levy.
\newblock Pick-a-pic: an open dataset of user preferences for text-to-image generation.
\newblock In \emph{Proceedings of the 37th International Conference on Neural Information Processing Systems}, Red Hook, NY, USA, 2023. Curran Associates Inc.

\bibitem[Li et~al.(2022)Li, Li, Xiong, and Hoi]{blip}
Junnan Li, Dongxu Li, Caiming Xiong, and Steven Hoi.
\newblock Blip: Bootstrapping language-image pre-training for unified vision-language understanding and generation.
\newblock In \emph{ICML}, 2022.

\bibitem[Lin et~al.(2023)Lin, Gao, Tang, Takikawa, Zeng, Huang, Kreis, Fidler, Liu, and Lin]{magic3d}
Chen-Hsuan Lin, Jun Gao, Luming Tang, Towaki Takikawa, Xiaohui Zeng, Xun Huang, Karsten Kreis, Sanja Fidler, Ming-Yu Liu, and Tsung-Yi Lin.
\newblock Magic3d: High-resolution text-to-3d content creation.
\newblock In \emph{Proceedings of the IEEE/CVF Conference on Computer Vision and Pattern Recognition (CVPR)}, pages 300--309, 2023.

\bibitem[Liu et~al.(2023{\natexlab{a}})Liu, Li, Wu, and Lee]{llava}
Haotian Liu, Chunyuan Li, Qingyang Wu, and Yong~Jae Lee.
\newblock Visual instruction tuning, 2023{\natexlab{a}}.

\bibitem[Liu et~al.(2023{\natexlab{b}})Liu, Shi, Chen, Zhang, Xu, Wei, Chen, Zeng, Gu, and Su]{zero12345++}
Minghua Liu, Ruoxi Shi, Linghao Chen, Zhuoyang Zhang, Chao Xu, Xinyue Wei, Hansheng Chen, Chong Zeng, Jiayuan Gu, and Hao Su.
\newblock One-2-3-45++: Fast single image to 3d objects with consistent multi-view generation and 3d diffusion.
\newblock \emph{arXiv preprint arXiv:2311.07885}, 2023{\natexlab{b}}.

\bibitem[Liu et~al.(2023{\natexlab{c}})Liu, Lin, Zeng, Long, Liu, Komura, and Wang]{syncdreamer}
Yuan Liu, Cheng Lin, Zijiao Zeng, Xiaoxiao Long, Lingjie Liu, Taku Komura, and Wenping Wang.
\newblock Syncdreamer: Generating multiview-consistent images from a single-view image.
\newblock \emph{arXiv preprint arXiv:2309.03453}, 2023{\natexlab{c}}.

\bibitem[Liu et~al.(2023{\natexlab{d}})Liu, Xie, Liu, and Wong]{syncmvd}
Yuxin Liu, Minshan Xie, Hanyuan Liu, and Tien-Tsin Wong.
\newblock Text-guided texturing by synchronized multi-view diffusion.
\newblock \emph{arXiv preprint arXiv:2311.12891}, 2023{\natexlab{d}}.

\bibitem[Melas-Kyriazi et~al.(2024)Melas-Kyriazi, Laina, Rupprecht, Neverova, Vedaldi, Gafni, and Kokkinos]{im3d}
Luke Melas-Kyriazi, Iro Laina, Christian Rupprecht, Natalia Neverova, Andrea Vedaldi, Oran Gafni, and Filippos Kokkinos.
\newblock Im-3d: Iterative multiview diffusion and reconstruction for high-quality 3d generation.
\newblock \emph{arXiv preprint arXiv:2402.08682}, 2024.

\bibitem[Meshy(2024)]{meshy}
Meshy.
\newblock Meshy text-to-{3D} v3.0, 2024.

\bibitem[Metzer et~al.(2022)Metzer, Richardson, Patashnik, Giryes, and Cohen-Or]{latentnerf}
Gal Metzer, Elad Richardson, Or Patashnik, Raja Giryes, and Daniel Cohen-Or.
\newblock Latent-nerf for shape-guided generation of 3d shapes and textures.
\newblock \emph{arXiv preprint arXiv:2211.07600}, 2022.

\bibitem[Mildenhall et~al.(2020)Mildenhall, Srinivasan, Tancik, Barron, Ramamoorthi, and Ng]{Nerf}
Ben Mildenhall, Pratul~P. Srinivasan, Matthew Tancik, Jonathan~T. Barron, Ravi Ramamoorthi, and Ren Ng.
\newblock Nerf: Representing scenes as neural radiance fields for view synthesis.
\newblock In \emph{Computer Vision – ECCV 2020: 16th European Conference, Glasgow, UK, August 23–28, 2020, Proceedings, Part I}, page 405–421, Berlin, Heidelberg, 2020. Springer-Verlag.

\bibitem[Niemeyer et~al.(2020)Niemeyer, Mescheder, Oechsle, and Geiger]{occupancy_fields}
Michael Niemeyer, Lars Mescheder, Michael Oechsle, and Andreas Geiger.
\newblock Differentiable volumetric rendering: Learning implicit 3d representations without 3d supervision.
\newblock In \emph{2020 IEEE/CVF Conference on Computer Vision and Pattern Recognition (CVPR 2020)}, pages 3501 -- 3512, Piscataway, NJ, 2020. IEEE.

\bibitem[OpenAI(2023)]{GPT4}
OpenAI.
\newblock Gpt-4 technical report.
\newblock 2023.

\bibitem[Oquab et~al.(2023)Oquab, Darcet, Moutakanni, Vo, Szafraniec, Khalidov, Fernandez, Haziza, Massa, El-Nouby, Assran, Ballas, Galuba, Howes, Huang, Li, Misra, Rabbat, Sharma, Synnaeve, Xu, J{\'e}gou, Mairal, Labatut, Joulin, and Bojanowski]{dinov2}
Maxime Oquab, Timoth{\'e}e Darcet, Th{\'e}o Moutakanni, Huy~Q. Vo, Marc Szafraniec, Vasil Khalidov, Pierre Fernandez, Daniel Haziza, Francisco Massa, Alaaeldin El-Nouby, Mahmoud Assran, Nicolas Ballas, Wojciech Galuba, Russ Howes, Po-Yao~(Bernie) Huang, Shang-Wen Li, Ishan Misra, Michael~G. Rabbat, Vasu Sharma, Gabriel Synnaeve, Huijiao Xu, Herv{\'e} J{\'e}gou, Julien Mairal, Patrick Labatut, Armand Joulin, and Piotr Bojanowski.
\newblock Dinov2: Learning robust visual features without supervision.
\newblock \emph{ArXiv}, abs/2304.07193, 2023.

\bibitem[Poole et~al.(2022)Poole, Jain, Barron, and Mildenhall]{dreamfusion}
Ben Poole, Ajay Jain, Jonathan~T. Barron, and Ben Mildenhall.
\newblock Dreamfusion: Text-to-3d using 2d diffusion.
\newblock \emph{arXiv}, 2022.

\bibitem[Qian et~al.(2024)Qian, Mai, Hamdi, Ren, Siarohin, Li, Lee, Skorokhodov, Wonka, Tulyakov, and Ghanem]{Magic123}
Guocheng Qian, Jinjie Mai, Abdullah Hamdi, Jian Ren, Aliaksandr Siarohin, Bing Li, Hsin-Ying Lee, Ivan Skorokhodov, Peter Wonka, Sergey Tulyakov, and Bernard Ghanem.
\newblock Magic123: One image to high-quality 3d object generation using both 2d and 3d diffusion priors.
\newblock In \emph{The Twelfth International Conference on Learning Representations (ICLR)}, 2024.

\bibitem[Radford et~al.(2021)Radford, Kim, Hallacy, Ramesh, Goh, Agarwal, Sastry, Askell, Mishkin, Clark, Krueger, and Sutskever]{clip_radford}
Alec Radford, Jong~Wook Kim, Chris Hallacy, Aditya Ramesh, Gabriel Goh, Sandhini Agarwal, Girish Sastry, Amanda Askell, Pamela Mishkin, Jack Clark, Gretchen Krueger, and Ilya Sutskever.
\newblock Learning transferable visual models from natural language supervision, 2021.

\bibitem[Rombach et~al.(2021)Rombach, Blattmann, Lorenz, Esser, and Ommer]{LatentDiffusion}
Robin Rombach, Andreas Blattmann, Dominik Lorenz, Patrick Esser, and Björn Ommer.
\newblock High-resolution image synthesis with latent diffusion models, 2021.

\bibitem[Saharia et~al.(2024)Saharia, Chan, Saxena, Lit, Whang, Denton, Ghasemipour, Ayan, Mahdavi, Gontijo-Lopes, Salimans, Ho, Fleet, and Norouzi]{imagen}
Chitwan Saharia, William Chan, Saurabh Saxena, Lala Lit, Jay Whang, Emily Denton, Seyed Kamyar~Seyed Ghasemipour, Burcu~Karagol Ayan, S.~Sara Mahdavi, Raphael Gontijo-Lopes, Tim Salimans, Jonathan Ho, David~J Fleet, and Mohammad Norouzi.
\newblock Photorealistic text-to-image diffusion models with deep language understanding.
\newblock In \emph{Proceedings of the 36th International Conference on Neural Information Processing Systems}, Red Hook, NY, USA, 2024. Curran Associates Inc.

\bibitem[Shi et~al.(2023{\natexlab{a}})Shi, Chen, Zhang, Liu, Xu, Wei, Chen, Zeng, and Su]{zero123plus}
Ruoxi Shi, Hansheng Chen, Zhuoyang Zhang, Minghua Liu, Chao Xu, Xinyue Wei, Linghao Chen, Chong Zeng, and Hao Su.
\newblock Zero123++: a single image to consistent multi-view diffusion base model, 2023{\natexlab{a}}.

\bibitem[Shi et~al.(2023{\natexlab{b}})Shi, Wang, Ye, Mai, Li, and Yang]{mvdream}
Yichun Shi, Peng Wang, Jianglong Ye, Long Mai, Kejie Li, and Xiao Yang.
\newblock Mvdream: Multi-view diffusion for 3d generation.
\newblock \emph{arXiv:2308.16512}, 2023{\natexlab{b}}.

\bibitem[Siddiqui et~al.(2024)Siddiqui, Monnier, Kokkinos, Kariya, Kleiman, Garreau, Gafni, Neverova, Vedaldi, Shapovalov, and Novotny]{assetgen}
Yawar Siddiqui, Tom Monnier, Filippos Kokkinos, Mahendra Kariya, Yanir Kleiman, Emilien Garreau, Oran Gafni, Natalia Neverova, Andrea Vedaldi, Roman Shapovalov, and David Novotny.
\newblock Meta 3d assetgen: Text-to-mesh generation with high-quality geometry, texture, and pbr materials.
\newblock \emph{arXiv}, 2024.

\bibitem[Tang et~al.(2023)Tang, Ren, Zhou, Liu, and Zeng]{dreamgaussian}
Jiaxiang Tang, Jiawei Ren, Hang Zhou, Ziwei Liu, and Gang Zeng.
\newblock Dreamgaussian: Generative gaussian splatting for efficient 3d content creation.
\newblock \emph{arXiv preprint arXiv:2309.16653}, 2023.

\bibitem[Team et~al.(2023)Team, Anil, Borgeaud, Alayrac, Yu, Soricut, Schalkwyk, Dai, Hauth, Millican, et~al.]{gemini}
Gemini Team, Rohan Anil, Sebastian Borgeaud, Jean-Baptiste Alayrac, Jiahui Yu, Radu Soricut, Johan Schalkwyk, Andrew~M Dai, Anja Hauth, Katie Millican, et~al.
\newblock Gemini: a family of highly capable multimodal models.
\newblock \emph{arXiv preprint arXiv:2312.11805}, 2023.

\bibitem[Thakur et~al.(2021)Thakur, Reimers, Daxenberger, and Gurevych]{SBERT}
Nandan Thakur, Nils Reimers, Johannes Daxenberger, and Iryna Gurevych.
\newblock Augmented {SBERT}: Data augmentation method for improving bi-encoders for pairwise sentence scoring tasks.
\newblock In \emph{Proceedings of the 2021 Conference of the North American Chapter of the Association for Computational Linguistics: Human Language Technologies}, pages 296--310, Online, 2021. Association for Computational Linguistics.

\bibitem[Tochilkin et~al.(2024)Tochilkin, Pankratz, Liu, Huang, Letts, Li, Liang, Laforte, Jampani, and Cao]{triposr}
Dmitry Tochilkin, David Pankratz, Zexiang Liu, Zixuan Huang, Adam Letts, Yangguang Li, Ding Liang, Christian Laforte, Varun Jampani, and Yan-Pei Cao.
\newblock Triposr: Fast 3d object reconstruction from a single image.
\newblock \emph{arXiv preprint arXiv:2403.02151}, 2024.

\bibitem[Touvron et~al.(2023)Touvron, Lavril, Izacard, Martinet, Lachaux, Lacroix, Rozière, Goyal, Hambro, Azhar, Rodriguez, Joulin, Grave, and Lample]{llama}
Hugo Touvron, Thibaut Lavril, Gautier Izacard, Xavier Martinet, Marie-Anne Lachaux, Timothée Lacroix, Baptiste Rozière, Naman Goyal, Eric Hambro, Faisal Azhar, Aurelien Rodriguez, Armand Joulin, Edouard Grave, and Guillaume Lample.
\newblock Llama: Open and efficient foundation language models, 2023.

\bibitem[Wang et~al.(2022)Wang, Du, Li, Yeh, and Shakhnarovich]{scorejacobianchaining}
Haochen Wang, Xiaodan Du, Jiahao Li, Raymond~A. Yeh, and Greg Shakhnarovich.
\newblock Score jacobian chaining: Lifting pretrained 2d diffusion models for 3d generation.
\newblock \emph{arXiv preprint arXiv:2212.00774}, 2022.

\bibitem[Wang and Shi(2023)]{imagedream}
Peng Wang and Yichun Shi.
\newblock Imagedream: Image-prompt multi-view diffusion for 3d generation.
\newblock \emph{arXiv preprint arXiv:2312.02201}, 2023.

\bibitem[Wang et~al.(2004)Wang, Bovik, Sheikh, and Simoncelli]{ssim}
Zhou Wang, A.C. Bovik, H.R. Sheikh, and E.P. Simoncelli.
\newblock Image quality assessment: from error visibility to structural similarity.
\newblock \emph{IEEE Transactions on Image Processing}, 13\penalty0 (4):\penalty0 600--612, 2004.

\bibitem[Wang et~al.(2024)Wang, Lu, Wang, Bao, Li, Su, and Zhu]{prolific_dreamer}
Zhengyi Wang, Cheng Lu, Yikai Wang, Fan Bao, Chongxuan Li, Hang Su, and Jun Zhu.
\newblock Prolificdreamer: high-fidelity and diverse text-to-3d generation with variational score distillation.
\newblock In \emph{Proceedings of the 37th International Conference on Neural Information Processing Systems}, Red Hook, NY, USA, 2024. Curran Associates Inc.

\bibitem[Wu et~al.(2024)Wu, Yang, Li, Zhang, Liu, Guibas, Lin, and Wetzstein]{Gpt4vEval}
Tong Wu, Guandao Yang, Zhibing Li, Kai Zhang, Ziwei Liu, Leonidas Guibas, Dahua Lin, and Gordon Wetzstein.
\newblock Gpt-4v (ision) is a human-aligned evaluator for text-to-3d generation.
\newblock In \emph{Proceedings of the IEEE/CVF Conference on Computer Vision and Pattern Recognition}, pages 22227--22238, 2024.

\bibitem[Xiang et~al.(2024)Xiang, Lv, Xu, Deng, Wang, Zhang, Chen, Tong, and Yang]{trellis}
Jianfeng Xiang, Zelong Lv, Sicheng Xu, Yu Deng, Ruicheng Wang, Bowen Zhang, Dong Chen, Xin Tong, and Jiaolong Yang.
\newblock Structured 3d latents for scalable and versatile 3d generation.
\newblock \emph{arXiv preprint arXiv:2412.01506}, 2024.

\bibitem[Xu et~al.(2023)Xu, Liu, Wu, Tong, Li, Ding, Tang, and Dong]{imagereward}
Jiazheng Xu, Xiao Liu, Yuchen Wu, Yuxuan Tong, Qinkai Li, Ming Ding, Jie Tang, and Yuxiao Dong.
\newblock Imagereward: learning and evaluating human preferences for text-to-image generation.
\newblock In \emph{Proceedings of the 37th International Conference on Neural Information Processing Systems}, Red Hook, NY, USA, 2023. Curran Associates Inc.

\bibitem[Yariv et~al.(2020)Yariv, Kasten, Moran, Galun, Atzmon, Ronen, and Lipman]{sdf}
Lior Yariv, Yoni Kasten, Dror Moran, Meirav Galun, Matan Atzmon, Basri Ronen, and Yaron Lipman.
\newblock Multiview neural surface reconstruction by disentangling geometry and appearance.
\newblock \emph{Advances in Neural Information Processing Systems}, 33, 2020.

\bibitem[Yi et~al.(2024)Yi, Fang, Wang, Wu, Xie, Zhang, Liu, Tian, and Wang]{gaussiandreamer}
Taoran Yi, Jiemin Fang, Junjie Wang, Guanjun Wu, Lingxi Xie, Xiaopeng Zhang, Wenyu Liu, Qi Tian, and Xinggang Wang.
\newblock Gaussiandreamer: Fast generation from text to 3d gaussians by bridging 2d and 3d diffusion models.
\newblock In \emph{CVPR}, 2024.

\bibitem[Yue et~al.(2024)Yue, Das, Engelmann, Tang, and Lenssen]{fit3d}
Yuanwen Yue, Anurag Das, Francis Engelmann, Siyu Tang, and Jan~Eric Lenssen.
\newblock {Improving 2D Feature Representations by 3D-Aware Fine-Tuning}.
\newblock In \emph{European Conference on Computer Vision (ECCV)}, 2024.

\bibitem[Zhang et~al.(2018)Zhang, Isola, Efros, Shechtman, and Wang]{lpips}
Richard Zhang, Phillip Isola, Alexei~A Efros, Eli Shechtman, and Oliver Wang.
\newblock The unreasonable effectiveness of deep features as a perceptual metric.
\newblock In \emph{CVPR}, 2018.

\bibitem[Zhu et~al.(2023)Zhu, Zhuang, and Koyejo]{HIFAHT}
Junzhe Zhu, Peiye Zhuang, and Oluwasanmi Koyejo.
\newblock Hifa: High-fidelity text-to-3d generation with advanced diffusion guidance.
\newblock In \emph{International Conference on Learning Representations}, 2023.

\end{thebibliography}
